\documentclass[lettersize,journal]{IEEEtran}
\usepackage{algorithm, algorithmicx}
\usepackage{algpseudocode}
\usepackage{amsmath,amssymb,amsfonts}
\usepackage{array}
\usepackage[style=ieee]{biblatex}
\usepackage{graphicx}
\usepackage[hidelinks]{hyperref}
\usepackage{longtable}
\usepackage[caption=false,font=normalsize,labelfont=sf,textfont=sf]{subfig}
\usepackage{stfloats}
\usepackage{tabularx}
    \newcolumntype{M}[1]{>{\centering\arraybackslash}m{#1}}
    
\usepackage{textcomp}
\usepackage{url}
\usepackage{verbatim}
\usepackage{xcolor}

\newcommand{\cvec}[1]{\boldsymbol{#1}}
\newcommand{\cvecdot}[1]{\dot{\boldsymbol{#1}}}
\newcommand{\cvecddot}[1]{\ddot{\boldsymbol{#1}}}
\newcommand{\cvecbarup}[1]{\overline{\cvec{#1}}}
\newcommand{\cvecbardown}[1]{\underline{\cvec{#1}}}
\newcommand{\cmat}[1]{\boldsymbol{#1}}
\newcommand{\cmatdot}[1]{\dot{\boldsymbol{#1}}}
\newcommand{\bracketref}[1]{(\ref{#1})}

\DeclareRobustCommand{\rchi}{{\mathpalette\irchi\relax}}
\newcommand{\irchi}[2]{\raisebox{\depth}{$#1\chi$}}

\begin{document}


\title{A General Framework for Hierarchical Redundancy Resolution Under Arbitrary Constraints*}

\author{Mario D. Fiore$^{1,2}$, Gaetano Meli$^{2,3}$, Anton Ziese$^{4}$, Bruno Siciliano$^{3}$ and Ciro Natale$^{1}$ 
\thanks{$^{1}$Authors are with the Dipartimento di Ingegneria, Università della Campania "Luigi Vanvitelli", 81031 Aversa, IT.}%
\thanks{$^{2}$Authors are with the Technology \& Innovation Center of KUKA Deutschland GmbH, 86165 Augsburg, DE.}%
\thanks{$^{3}$Authors are with the Department of Electrical Engineering and Information Technology, University of Naples Federico II, 80125 Naples, IT.}%
\thanks{$^{4}$Authors are with the TU Darmstadt, 64289 Darmstadt, DE}%
\thanks{This work was partly supported by the German Federal Ministry of Education and Research (BMBF) through the project Internet of Construction (grant no. 02P17D083).}%
\thanks{*This work has been submitted to the IEEE for possible publication.
Copyright may be transferred without notice, after which this version may
no longer be accessible}}



\maketitle

\begin{abstract}
The increasing interest in autonomous robots with a high number of degrees of freedom for industrial applications and service robotics demands  control  algorithms  to  handle  multiple tasks  as  well  as  hard constraints  efficiently. This paper presents a general framework in which both kinematic (velocity- or acceleration-based) and dynamic (torque-based) control of redundant robots are handled in a unified fashion. The framework allows for the specification of redundancy resolution problems featuring a hierarchy of arbitrary (equality and inequality) constraints, arbitrary weighting of the control effort in the cost function and an additional input used to optimize possibly remaining redundancy. To solve such problems, a generalization of the Saturation in the Null Space (SNS) algorithm is introduced, which extends the original method according to the features required by our general control framework. Variants of the developed algorithm are presented, which ensure both efficient computation and optimality of the solution. Experiments on a KUKA LBRiiwa robotic arm, as well as simulations with a highly redundant mobile manipulator are reported.
\end{abstract}

\begin{IEEEkeywords}
Redundant Robots, Motion Control, Optimization and Optimal Control, Hierarchical Control, Redundancy Resolution.
\end{IEEEkeywords}

\section{Introduction}
\label{sec:1intro}
\IEEEPARstart{R}{obots} with a large number of degrees of freedom (DOFs) have seen an increasing demand in industrial applications and service robotics in recent years. From industrial manipulators \cite{scheurer2016industrial} up to humanoid robots \cite{escande2014hierarchical} and unmanned vehicles \cite{antonelli2014underwater, baizid2015experiments}, robots are more and more often designed to be highly redundant, i.e., to have significantly more DOFs than are needed to perform a given task. Redundancy allows for more versatility in performing an assigned task and for the simultaneous specification of multiple objectives. However, proper redundancy resolution strategies are required to compute suitable joint motions and achieve effective exploitation of the redundant DOFs. Tasks can consist of equality (e.g., tracking of a Cartesian trajectory) or inequality (e.g., keeping a joint within its allowed range of motion) constraints and present different priority based on their respective importance. For example, safety-related tasks, such as avoiding collision with a human or the environment, may be considered of utmost importance and therefore assigned the highest priority. On the other hand, operational tasks, such as following a desired end-effector trajectory, may be handled with lower priority and carried out only if their execution does not interfere with the fulfilment of higher-priority tasks \cite{scheurer2016industrial, di2021framework}. Furthermore, since the robots are often required to autonomously operate in unstructured environments, redundancy resolution algorithms must be able to find a solution for the specified tasks online. Moreover, one or more tasks may become infeasible altogether due to some change in the environment, in which case the robot must retain robust and predictable behavior, eventually leading to a controlled stop, if necessary. Depending on the specific application and the actual data availability, the redundancy resolution algorithm may make use of the robot differential kinematic equations or dynamic model. Finally, when considering robots with a high number of DOFs, computational efficiency becomes considerably important in light of real-time capability. 

\subsection{Relevant Literature}
Redundancy resolution methods have been extensively studied for decades. Using the first-order differential kinematic model of the robot, \cite{Whitney.1969} first introduced a method for solving a single-task problem based on a minimum-norm solution. A task priority approach exploiting null space projection was then presented in \cite{maciejewski1985obstacle} for two tasks and later extended to a generic number of tasks in \cite{Siciliano.1991}. Similar approaches have been developed at acceleration level, using the second-order differential kinematic model \cite{Siciliano.1992}. The exploitation of the robot dynamical model has instead led to the Operation Space Formulation \cite{khatib1983dynamic} and its extension to prioritized tasks \cite{sentis2004prioritized, ott2015prioritized, dietrich2019hierarchical}.

In all the above-mentioned contributions, the tasks assigned to the robotic system consist of a set of equality constraints. However, several tasks may be naturally described as inequality constraints. Two common, yet very important, examples are joint limits and collision avoidance. Nevertheless, other cases can easily be found, like the handling of virtual walls, the avoidance of visibility loss or the control of the center of mass in legged robots. Handling of inequality constraints has been tackled in different ways. Most methods are based on pseudoinversion of task Jacobian matrices and null space projections, and {\color{black} here referred to as \textit{analytical}}. An early approach in this class of methods is represented by the Gradient-Based Projection (GBP) method \cite{liegeois1977automatic}, in which inequalities are converted into a cost function. Robot redundancy is then exploited to minimize such cost, in the attempt of keeping the task variables within their allowed range. However, the optimization is performed at a lower priority level, meaning that fulfilment of the inequality constraints is not guaranteed. Another simple, yet effective, approach is to perform task scaling \cite{hollerbach1983dynamic, chiacchio1995coping, antonelli2003new}, i.e., to reduce the speed (or acceleration) required by equality constraint task commands to recover feasibility with respect to one or more violated inequality constraints. One benefit of the task scaling method is that the directions of the equality constraints task commands are preserved. The execution of such commands is only extended in time, which is in many cases acceptable. Furthermore, task scaling can easily be applied to the case of prioritized tasks and provides the robot with a predictable behavior in case of conflict between equality and inequality constraints. On the other hand, the mere use of this technique implies slower task execution every time an inequality constraint would be violated. Before resorting to such a drastic measure, one should first verify that no alternative joint motions exist, which can ensure the satisfaction of both equality and inequality constraints.

Another widely adopted approach is to convert inequality constraints into equivalent equality constraints. To avoid overconstrained motions, these additional equality constraints can only be added to the set of prioritized tasks (often referred as \textit{Stack of Tasks}, SoT) with the lowest priority. Thus, the satisfaction of the original inequalities is not guaranteed \cite{Khatib.1986}. Alternatively, the satisfaction of inequality constraints can be monitored and the insertion of equivalent equality constraints only carried out at the desired priority level when task inequality bounds are approached. A classic method in torque-based redundancy resolution is to resort to artificial potential fields \cite{sentis2005synthesis}, in which virtual forces are used to push task variables away from the bounds of their allowed range. The effectiveness of this method, however, strongly depends on the parameters that regulate the activation and the intensity of the repulsive virtual force \cite{munoz2018operational}. Using a similar approach, methods for handling inequality constraints in velocity-based schemes were developed in \cite{mansard2009unified, moe2016set}. Also in this case, special parameters are introduced to define activation functions and thresholds, as well as to shape the functions that force task variables to stay in a \textit{safe} range. Furthermore, handling the continuous activation/deactivation of the additional equality constraints may require parallel computation of several possible solutions \cite{moe2016set, di2021framework}. Additionally, online modifications to the SoT typically produce discontinuities in the solution \cite{sentis2005synthesis}. Besides requiring more parametrization, methods to ensure smooth insertion to/removal from the SoT typically fail at respecting the strict priority among the tasks during transitions \cite{mansard2009unified} or imply a significant increase of the computational cost \cite{lee2012intermediate, liu2016generalized}. 

Given the above-mentioned limitations of analytical approaches in dealing with inequality constraints, numerical methods based on Hierarchical Quadratic Programming (HQP) have been more intensively investigated in recent years \cite{kanoun2011kinematic, escande2014hierarchical, aertbelien2014etasl, liu2016generalized, hoffman2018multi, quiroz2019whole}. The main idea, first introduced in \cite{kanoun2011kinematic}, is to solve a cascade of QP problems, one for each level of priority. Both equality and inequality constraints can be easily specified in the context of QP at each priority level. To enforce strict priority among tasks, each QP problem should not increase the minimum obtained by solving the previous problem. Although effective in fulfilling the given constraints, the technique suffered from a high computational cost, when compared to analytical solutions. This effect has been later mitigated in \cite{escande2014hierarchical,liu2016generalized} by formalizing the multiple QP problems in a single one and using complete orthogonal decomposition to obtain the null spaces of the prioritized tasks. Another drawback of the mentioned solutions based on HQP is that they do not feature a defined behavior in case a task is not feasible. In fact, infeasible tasks are normally handled through the introduction of slack variables and/or by relaxing the constraints in a least square sense. 
{\color{black}However, such strategy does not deliver a predictable robot behavior. On the other hand, in many robot applications (especially in the industrial field) it is acceptable to just extend the execution of the equality constraints over time, while preserving the task directions. In all these cases, task scaling seems a more appropriate solution to recover feasibility.}

The \textit{Saturation in the Null Space} (SNS) algorithm \cite{flacco2015control} for kinematic control provides a promising link between analytical methods and HQP, while featuring a task scaling technique. Task prioritization uses null space projections as in analytical methods. At each level of priority, inequality constraints are monitored and, if necessary, converted into equality constraints in a similar way as in (active-set) QP. Finally, a task scaling strategy is applied in all cases where it was not possible to find a feasible solution. Although special variants of this algorithm tackle additional important aspects like the optimality of the solution and the efficiency of the computation, several points still remain unsolved. First, the algorithm in \cite{flacco2015control} only handles joint space inequality constraints, which are always treated with the highest priority. Thus, operational inequality constraints, defined in task space and/or having lower priority are not directly addressed. Moreover, the optimization process presents a simple cost function, based on the idea of minimizing the pure joint velocity norm. Similarly to \cite{escande2014hierarchical}, also the machinery used to speed-up the computation strongly relies on the consideration of such cost function. However, joint velocity minimization has been shown to have drawbacks for robots whose joint coordinates present mixed units, which is often the case when considering robots with a large number of DOFs, e.g., mobile manipulators \cite{lachner2020influence}. This aspect is also essential in the context of energy-aware motion generation \cite{lachner2021energy}. Therefore, a proper redundancy resolution algorithm should provide the possibility to select more appropriate metrics.

\subsection{Contribution and Structure of the Paper}
Most of the above-mentioned solutions utilize velocity-based redundancy resolution, i.e., the robot is described using a first order kinematic equation, mostly because of its mathematical simplicity. However, moving to second-order algorithms, i.e., acceleration or torque level, offers some advantages, e.g., enabling the inequality constraints to  include maximum/minimum accelerations, and also improving the noise, vibration and harshness (NVH) behavior of the robot. Torque-based approaches additionally offer compliance, allowing the robot to safely handle physical contact. On the other hand, these methods suffer from dynamic model uncertainties as discussed in \cite{di2021effects} and as evidenced by some of the experimental results presented in Sect. \ref{sec:4_results_exp_iiwa}.

This paper presents a general framework for hierarchical redundancy resolution under arbitrary (equality and inequality) constraints, consolidating and expanding the results of our recent work \cite{ziese2020redundancy}. A general description of the redundancy resolution problem is proposed, which allows bringing all of the aforementioned redundancy resolution schemes (velocity-, acceleration- and torque-based) in a unified form. Building on \cite{ziese2020redundancy}, the proposed generalized redundancy resolution problem is further developed to specifically include arbitrary equality and inequality constraints on every priority level. Furthermore, an additional input has been considered, which allows for the specification of optimization criteria to better manage possible residual redundant DOFs. Having a unified description of the redundancy resolution problem also allows for the design of a single solver. Here, we introduce a novel algorithm, {\color{black}named \textit{extended SNS} (\textit{eSNS})}, which builds on the SNS methodology extending it in many significant aspects. Being capable of solving the proposed generalized redundancy resolution problem, the algorithm can indistinctly handle velocity-, acceleration- or torque-based schemes. Moreover, arbitrary inequality constraints can be managed at each level of priority. Finally, arbitrary metrics and additional inputs are considered in the optimization process when computing a solution. 

An analysis of the computational efficiency is also carried out, leading to a special variant of the algorithm, {\color{black}named \textit{Fast-eSNS}, which} directly extends the work in \cite{ziese2020redundancy}. Another variant {\color{black}(\textit{Opt-eSNS})} can be obtained as a result of the analysis on the optimality of the solution and is also introduced in this paper. Finally, a novel shaping of the inequality constraint bounds is proposed, which allows treating position, velocity and acceleration limitations in a unified way. The parametrization of such shaping is consistent with the one typically used to define task commands in equality constraints. Thus, it is based on the desired behaviour of dynamic systems and embeds a strong physical meaning. It is also shown that such shaping represents an extension of the solutions proposed in \cite{flacco2015control, flacco2012prioritized} for the SNS algorithm and \cite{osorio2019physical}.

This paper is organized as follows. Section \ref{sec:2_genredres} introduces the formalism and the mathematical background on redundancy resolution. Furthermore, the proposed shaping is introduced for both task references (equality constraints) and bounds (inequality constraints). Lastly, the generalized redundancy resolution problem is presented. Section \ref{sec:3_extendedSNS} introduces the proposed redundancy resolution algorithm, its relation to the original SNS method and the two additional variants developed in light of computational efficiency and optimality of the solution. Finally, simulation and experimental results are reported in Sect. \ref{sec:4_results} using a KUKA LBRiiwa robot and a mobile dual-arm manipulator to prove the effectiveness and evaluate the performance of the proposed algorithms.

\section{Generalized Redundancy Resolution}
\label{sec:2_genredres}
As mentioned in the introduction, there are different approaches to control redundant robots depending on the model used to describe the robot behavior. Velocity- \cite{Whitney.1969} and acceleration-based \cite{Siciliano.1992} models describe the robot using a kinematic equation at velocity and acceleration level, respectively; the output of the control algorithm is then a vector of joint velocities or accelerations, which are typically integrated to compute joint positions that are then given as input to underlying joint position controllers. A torque-based approach \cite{khatib1983dynamic}, as the name suggests, outputs a vector of joint torques instead. These are then directly sent to joint actuators and an actual control loop is closed using measured joint position and velocity.

The basics of all the above-mentioned redundancy resolution schemes are briefly recalled in Sect. \ref{sec:2a_mathback}. Section \ref{sec:2b_shaping} discusses the shaping of the task commands for both equality and inequality constraints. The reformulation of all redundancy resolution models and constraints as one generalized optimization problem is then presented in Sect. \ref{sec:2c_genProb}.

\subsection{Mathematical background}
\label{sec:2a_mathback}
Let $\cvec{x} \in \mathcal{X}$ be the vector of variables describing a $m$-dimensional task, with $\mathcal{X}$ being a domain of $\mathbb{R}^m$. Moreover, let $\cvec{q} \in \mathcal{Q} \subseteq \mathbb{R}^n$ be the vector of the coordinates describing the joint configuration of a given $n$-DOFs robot, with $n \geq m$. The relation between $\cvec{q}$ and $\cvec{x}$ is given by the forward task kinematics equation
\begin{equation*}
    \label{eq:fk}
	\cvec{f}: \cvec{q} \in \mathcal{Q} \subseteq \mathbb{R}^n \to \cvec{x} \in \mathcal{X} \subseteq \mathbb{R}^m ,
	\ \
	\cvec{x} = \cvec{f}(\cvec{q}),
\end{equation*}
while the first and second order differential task kinematics relations are given by
\begin{subequations}
    \label{eq:fkDiff}
    \begin{align}
    	\cvecdot{x}(t) &= \cmat{J}(\cvec{q}(t))\cvecdot{q}(t)    \label{eq:fkDiff_vel}  \\
    	\cvecddot{x}(t) &= \cmat{J}(\cvec{q}(t))\cvecddot{q}(t) +  \cmatdot{J}(\cvec{q}(t))\cvecdot{q}(t),                   \label{eq:fkDiff_acc}
    \end{align}
\end{subequations}
where $\cmat{J}(\cvec{q}(t))$ is the $m \times n$ task Jacobian matrix. Note that a task might be defined directly in the joint space, namely,
\begin{equation*}
    \label{eq:jointTask}
 	    \cvec{x}(t) = \cvec{q}(t) \ \Rightarrow  \ \cmat{J}(\cvec{q}) = \cmat{I} \qquad \qquad \forall \cvec{q} \in \mathcal{Q},
\end{equation*}
where $\cmat{I}$ is the identity matrix. This special case is often considered to track joint space references or to handle joint limitations, i.e., to ensure that all joint positions (as well as velocities and accelerations) stay within the allowed mechanical ranges. 

Velocity-based redundancy resolution methods typically operate an inversion of the differential map \bracketref{eq:fkDiff_vel} to ensure that the task velocity, $\cvecdot{x}(t)$, tracks a given reference, $\cvec{r}_v(t)$, or stays within a certain range characterized by lower bounds $\cvecbardown{r}_v(t)$ and upper bounds $\cvecbarup{r}_v(t)$. Thus, joint velocities are computed so as to satisfy equality or inequality constraints in the form \begin{subequations}
    \label{eq:velconstr}
    \begin{align}
	    \cvecdot{x}(t) &= \cvec{r}_v(t), \label{eq:velconstr_eq} \\
	    \cvecbardown{r}_v(t) \leq &\cvecdot{x}(t) \leq \cvecbarup{r}_v(t). \label{eq:velconstr_ineq}
    \end{align}
\end{subequations}
where the inequalities are intended component-wise.

Similarly to velocity schemes, acceleration-based redundancy resolution methods operate an inversion of the second-order differential map \bracketref{eq:fkDiff_acc} to enforce equality or inequality constraints in the form
\begin{subequations}
    \label{eq:accconstr}
    \begin{align}
	    \cvecddot{x}(t) &= \cvec{r}_a(t), \label{eq:accconstr_eq} \\
	    \cvecbardown{r}_a(t) \leq &\cvecddot{x}(t) \leq \cvecbarup{r}_a(t). \label{eq:accconstr_ineq}
    \end{align}
\end{subequations}

Finally, torque-based methods use the joint space dynamic model
\begin{equation}
    \label{eq:dynMod}
	\cmat{M}(\cvec{q}(t)) \cvecddot{q}(t) + \cvec{c}(\cvec{q}(t), \cvecdot{q}(t)) + \cvec{g}(\cvec{q}(t)) = \cvec{\tau}(t),
\end{equation}
where $\cmat{M}(\cvec{q}(t)) \in \mathbb{R}^{n \times n}$ is the positive definite robot inertia matrix, $\cvec{c}(\cvec{q}(t), \cvecdot{q}(t)) \in \mathbb{R}^n$ is the vector of joint space Coriolis and centrifugal forces, $\cvec{g}(\cvec{q}(t)) \in \mathbb{R}^n$ is the vector of the joint space gravity forces, and $\cvec{\tau}(t)$ represents the vector of generalized joint torques. {\color{black} It should be noticed that, for the scope of this work, no external forces acting on the robot are considered}. Isolating the joint acceleration $\cvecddot{q}(t)$ in \bracketref{eq:dynMod} and substituting it in \bracketref{eq:fkDiff_acc} yields
\begin{equation}
    \begin{split}
        \label{eq:fk_dynMod_trq}
    	\cvecddot{x}(t) = \cmat{J}(\cvec{q}(t)) &\cmat{M}^{-1}(\cvec{q}(t)) \big(\cvec{\tau}(t)-\cvec{c}(\cvec{q}(t), \cvecdot{q}(t)) - \cvec{g}(\cvec{q}(t))\big) \\
    	& + \cmatdot{J}(\cvec{q}(t))\cvecdot{q}(t),
    \end{split}
\end{equation}
which relates the task acceleration $\cvecddot{x}(t)$ to the joint torques $\cvec{\tau}(t)$. Starting from \bracketref{eq:fk_dynMod_trq}, constraints on the task acceleration $\cvecddot{x}(t)$ {\color{black} can be chosen for torque-based schemes as in \bracketref{eq:accconstr}. Moreover, constraints on the maximum torque could also be of interest for this class of methods. Such constraints can be expressed as: 
\begin{equation}
    \label{eq:trqconstr}
    \cvecbardown{\tau}(t) \leq  \cvec{\tau}(t) \leq \cvecbarup{\tau}(t).
\end{equation}
}

\subsection{Shaping of task references and bounds}
\label{sec:2b_shaping}
From the analysis in the previous section it can be easily recognized that velocity-based redundancy resolution methods aim at satisfying task space constraints defined at velocity level, while acceleration and torque schemes involve constraints on task space accelerations. However, it is often desired to satisfy constraints defined on position level, i.e., directly involving the vector of task variables $\cvec{x}(t)$. This requires a particular shaping of the references and of the bounds in \bracketref{eq:velconstr} and \bracketref{eq:accconstr}, according to the particular redundancy resolution scheme.

\begin{figure*}
    \centering
    \includegraphics[width=0.99\textwidth]{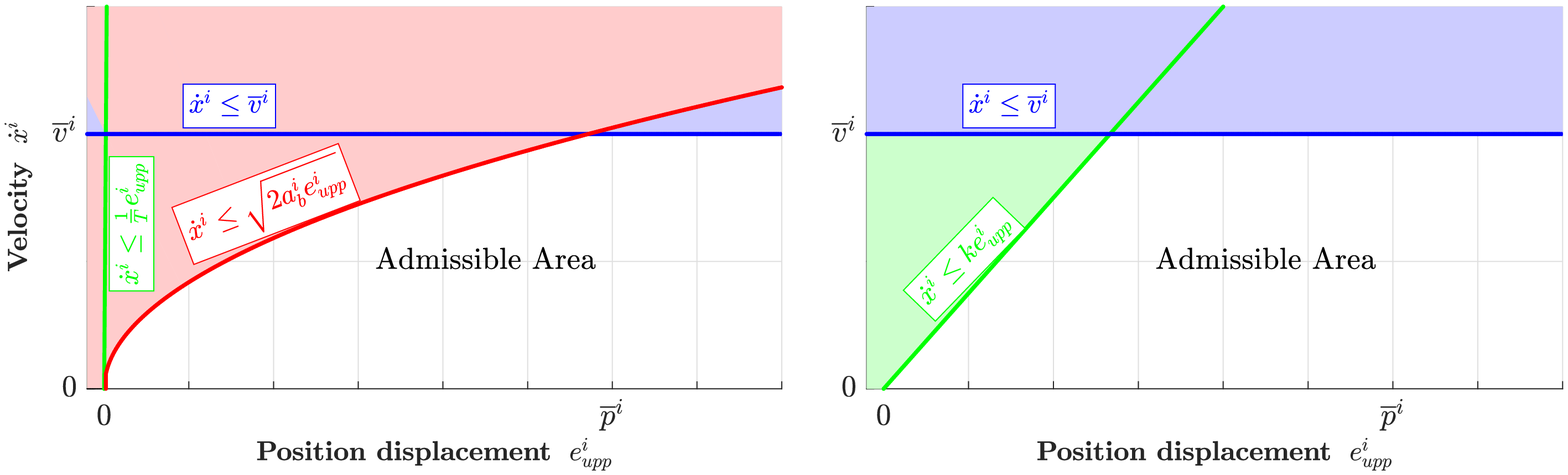}
\caption{Shaping of the upper velocity bound in case of $\dot{\overline{p}^i}=0$ and $\overline{v}^i(t) = \overline{v}^i$: the shaping obtained through the formulation in \bracketref{eq:velbounds_snsorig} is reported on the left, whereas the one obtained with the proposed shaping \bracketref{eq:velbounds} is reported on the right; the white areas represent the set of admissible pairs $\left( e^i_{upp}(t), \dot{x}^i(t) \right)$ according to the corresponding bound shaping. The figures have been generated using $T=0.005\,$s, $\overline{v}^i=2 \,$m/s, $\overline{p}^i=3 \,$m, $a^i_b=0.75 \,$m/s\textsuperscript{2} and $\cmat{K} = k\cmat{I}$, with $k=1.5 \,$s\textsuperscript{-1}.}
    \label{fig:boundsComparison}
\end{figure*}

Assume a desired task space trajectory to be defined by $(\cvec{x}_d(t), \cvecdot{x}_d(t), \cvecddot{x}_d(t))$. To ensure exponential convergence of $\cvec{x}(t)$ to $\cvec{x}_d(t)$, and to avoid numerical drift of the solution, the task reference velocity $\cvec{r}_v(t)$ is usually chosen in velocity-based schemes as \cite{balestrino1984robust} 
\begin{equation}
    \label{eq:clik_vel}
    \cvec{r}_v(t) = \cvecdot{x}_d(t) + \cmat{K}\cvec{e}(t),
\end{equation}
with $\cmat{K} \in \mathbb{R}^{m \times m}$ being a positive definite (p.d.) matrix and $\cvec{e}(t)$ the displacement between $\cvec{x}_d(t)$ and $\cvec{x}(t)$. The choice \bracketref{eq:clik_vel} imposes an evolution of task variables as a first order linear system with a convergence rate depending on the eigenvalues of $\cmat{K}$. Analogously, the task reference acceleration $\cvec{r}_a(t)$ in acceleration and torque schemes is typically chosen as \cite{khatib1983dynamic, Siciliano.1992}
\begin{equation}
    \label{eq:clik_acc}
    \cvec{r}_a(t) = \cvecddot{x}_d(t) + \cmat{D}\cvecdot{e}(t) + \cmat{K}\cvec{e}(t), \quad \cmat{D}, \cmat{K} \ \text{p.d.} \in \mathbb{R}^{m \times m}
\end{equation}
which imposes an evolution of the task variables as a second-order linear system.

In the case of inequality constraints, the following limits might instead exist on task space quantities:
\begin{subequations}
    \label{eq:posvelacclim}
    \begin{align}
	    \cvecbardown{p}(t) \leq \cvec{x}(t)     \leq \cvecbarup{p}(t) \label{eq:poslim}\\
	    \cvecbardown{v}(t) \leq \cvecdot{x}(t)  \leq \cvecbarup{v}(t) \label{eq:vellim}\\ 
	    \cvecbardown{a}(t) \leq \cvecddot{x}(t) \leq \cvecbarup{a}(t) \label{eq:acclim}
	\end{align}
\end{subequations}

To enforce the position limits \bracketref{eq:poslim} in velocity-based schemes, the lower and upper bounds $\cvecbardown{r}_v$ and $\cvecbarup{r}_v$ can be chosen according to \bracketref{eq:clik_vel}, namely \cite{aertbelien2014etasl}:
\begin{equation}
    \label{eq:velbounds_pos}
    \begin{split}
        \cvecbardown{r}_v(t) &= \dot{\cvecbardown{p}}(t) + \cmat{K}\cvec{e}_{low}(t) \\
        \cvecbarup{r}_v(t)   &= \dot{\cvecbarup{p}}(t) + \cmat{K}\cvec{e}_{upp}(t),
    \end{split}
\end{equation}
where $\cvec{e}_{low}$ is the displacement between $\cvecbardown{p}$ and $\cvec{x}$, and $\cvec{e}_{upp}$ the one between $\cvecbarup{p}$ and $\cvec{x}$. 
The effect of \bracketref{eq:velbounds_pos} is that the task variables are imposed to move towards their position limits no faster than a first order linear system with a convergence rate depending on the eigenvalues of $\cmat{K}$. The velocity limits \bracketref{eq:vellim} can be easily integrated in \bracketref{eq:velbounds_pos}: let $g^i$ denote the $i$th component of a vector $\cvec{g}$, and let ${\cvec{g}^i}^T$ indicate the row vector containing the $i$th row of a matrix $\cmat{G}$; then the $i$th components of $\cvecbardown{r}_v(t)$ and $\cvecbarup{r}_v(t)$ can be chosen as
\begin{equation}
    \label{eq:velbounds}
    \begin{split}
        \underline{r}^i_v(t) &= \max \left\{\dot{\underline{p}}^i(t) + {\cvec{k}^i}^T\cvec{e}_{low}(t), \ \underline{v}^i(t) \right\} \\
        \overline{r}^i_v(t)   &= \min \left\{\dot{\overline{p}}^i(t)+{\cvec{k}^i}^T\cvec{e}_{upp}(t), \  \overline{v}^i(t) \right\}.
    \end{split}
\end{equation}
The acceleration limits \bracketref{eq:acclim}, on the other hand, cannot be strictly enforced at velocity level. However, some techniques exist that tackle this problem \cite{rubrecht2012motion}.

It should be noticed that the proposed velocity bounds \bracketref{eq:velbounds} improve the shaping in \cite{flacco2015control}.
This can be written as
\begin{equation}
    \label{eq:velbounds_snsorig}
    \begin{split}
        \underline{r}^i_v(t) &= \max \left\{ \frac{e^i_{low}(t)}{T}, \ \underline{v}^i, -\sqrt{2a^i_b e^i_{low}(t)} \  \right\} \\
        \overline{r}^i_v(t)  &= \min \left\{ \frac{e^i_{upp}(t)}{T}, \ \overline{v}^i, \sqrt{2a^i_b e^i_{upp}(t)}   \right\},
    \end{split}
\end{equation}
with $T$ being the control cycle time and $a^i_b>0$ a parameter to adjust the shaping in the proximity of position limits. {\color{black}Ignoring for a moment the third term on the right-hand side of \bracketref{eq:velbounds_snsorig}}, it  can  be  easily recognized  that  the bounds proposed in \bracketref{eq:velbounds} generalize the ones in \bracketref{eq:velbounds_snsorig}. Indeed, the latter are obtained from \bracketref{eq:velbounds} assuming $\cvecbardown{p}$, $\cvecbarup{p}$, $\cvecbardown{v}$ and $\cvecbarup{v}$ to be constant over time and $\cvec{k}^i$ to be a vector having $1/T$ in the $i$th position and zero in all the remaining ones. Furthermore, the typically small value of $T$ (usually $1\div5$ milliseconds in nowadays controllers) allows task variables to have high speed in the proximity of position limits, leading to sudden deceleration when the limit is actually hit. This typically undesired effect is avoided by the introduction of the third term on the right-hand side of \bracketref{eq:velbounds_snsorig}, whose action can be regulated via the \textit{braking} parameter $a^i_b$. However, this makes the contribution of the first term on the right-hand side of \bracketref{eq:velbounds_snsorig} irrelevant in practice, as it always results in a less restricting bound for $\dot{x}^i(t)$, no matter the value of the displacement from the position limit. On the other hand, the proposed bounds \bracketref{eq:velbounds} allow, through the setting of ${\cvec{k}^i}^T$, to adjust the maximum rate of convergence (over time) of the task variable to the position limits and, thus, also to regulate the maximum allowed speed in the proximity of these limits. The above considerations can be better understood when looking at Fig. \ref{fig:boundsComparison}, in which a practical case of shaping of the upper velocity bound according to \bracketref{eq:velbounds} and \bracketref{eq:velbounds_snsorig} is presented.

Similarly to \bracketref{eq:velbounds_pos}, position limits can be enforced in acceleration- and torque-based schemes by choosing
\begin{equation}
    \label{eq:accbounds_pos}
    \begin{split}
        \cvecbardown{r}_a(t) &= \ddot{\cvecbardown{p}}(t) + \cmat{D}\cvecdot{e}_{low}(t) + \cmat{K}\cvec{e}_{low}(t) \\
        \cvecbarup{r}_a(t)   &= \ddot{\cvecbarup{p}}(t) + \cmat{D}\cvecdot{e}_{upp}(t) + \cmat{K}\cvec{e}_{upp}(t).
    \end{split}
\end{equation}
The  effect  of  \bracketref{eq:accbounds_pos}  is  that  the  task variables are imposed to move towards their position limits no faster than a second order linear system characterized by the matrices $\cmat{K}$ and $\cmat{D}$.
Recalling that $\cvecdot{e}_{low} = \dot{\cvecbardown{p}} - \cvecdot{x}$, $\cvecdot{e}_{upp} = \dot{\cvecbarup{p}} - \cvecdot{x}$ and introducing $\cmat{K}_1 = \cmat{D}^{-1}\cmat{K}$, the bounds \bracketref{eq:accbounds_pos} can be rewritten as
\begin{equation}
    \label{eq:accbounds_pos2}
    \begin{split}
        \cvecbardown{r}_a(t) &= \ddot{\cvecbardown{p}}(t) + \cmat{D}(\dot{\cvecbardown{p}}(t) + \cmat{K}_1\cvec{e}_{low}(t) - \cvecdot{x}(t)) \\
        \cvecbarup{r}_a(t)   &= \ddot{\cvecbarup{p}}(t) + \cmat{D}(\dot{\cvecbarup{p}}(t) + \cmat{K}_1\cvec{e}_{upp}(t) - \cvecdot{x}(t)).
    \end{split}
\end{equation}
Velocity limits \bracketref{eq:vellim} can then be integrated in \bracketref{eq:accbounds_pos2} by replacing the terms $\dot{\cvecbardown{p}}(t) + \cmat{K}_1\cvec{e}_{low}(t)$ and $\dot{\cvecbarup{p}}(t) + \cmat{K}_1\cvec{e}_{upp}(t)$ with the corresponding velocity bounds computed as in \bracketref{eq:velbounds}
\begin{equation*}
    \label{eq:accbounds_posvel}
    \begin{split}
        \cvecbardown{r}_a(t) &= \ddot{\cvecbardown{p}}(t) + \cmat{D}(\cvecbardown{r}_v(t) - \cvecdot{x}(t)) \\
        \cvecbarup{r}_a(t)   &= \ddot{\cvecbarup{p}}(t) + \cmat{D}(\cvecbarup{r}_v(t) - \cvecdot{x}(t)).
    \end{split}
\end{equation*}
Finally, the acceleration limits \bracketref{eq:acclim} can directly be integrated, leading to
\begin{equation}
    \label{eq:accbounds}
    \begin{split}
        \underline{r}^i_a(t) &= \max \left\{\ddot{\underline{p}}^i(t) + {\cvec{d}^i}^T(\underline{r}^i_v(t) - \cvecdot{x}(t)), \ \underline{a}^i(t) \right\} \\
        \overline{r}^i_a(t)   &= \min \left\{\ddot{\overline{p}}^i(t) + {\cvec{d}^i}^T(\overline{r}^i_v(t) - \cvecdot{x}(t)), \  \overline{a}^i(t) \right\}.
    \end{split}
\end{equation}
The proposed acceleration bounds \bracketref{eq:accbounds} generalize the shaping proposed in \cite{flacco2012prioritized, osorio2019physical}. This can be obtained by assuming constant (over time) position and velocity limits and by choosing $\cvec{d}^i$ as a vector having $1/T$ in the $i$th position and zero in all remaining ones. 

\subsection{Generalized control problem}
\label{sec:2c_genProb}
Consider the case in which, for a given task, a certain number of task variables are constrained by equality constraints, while the rest is subject to inequality constraints. Analogously, two different tasks might enforce equality and inequality constraints, separately. In such cases, the two types of constraint coexist for the given task specification.
Considering the models in \bracketref{eq:fkDiff} and \bracketref{eq:fk_dynMod_trq}, the constraints in \bracketref{eq:velconstr} and \bracketref{eq:accconstr} can all be brought into the form
\begin{subequations}
    \label{eq:genconstr}
    \begin{align}
	    \cmat{A} &\cvec{u} = \cvec{b} \\
	    \cvecbardown{d} \leq &\cmat{C} \cvec{u} \leq \cvecbarup{d}. \label{eq:genconstr_ineq}
	\end{align}
\end{subequations}
The definition of all the variables in \bracketref{eq:genconstr} for the different control schemes can be found in Tab. \ref{tab:unifiedControlProblem}, where the subscripts $eq$ and $iq$ are used to indicate the Jacobian matrices that are related to the task variables involved in equality and inequality constraints, respectively. For the sake of clarity, dependencies are omitted. {\color{black} Although not directly included in Tab. \ref{tab:unifiedControlProblem}, it should be noticed that the constraints \bracketref{eq:trqconstr} can also be brought to the form \bracketref{eq:genconstr_ineq} for torque-based schemes by setting $\cmat{C} = \cmat{I}$, $\cvecbardown{d} = \cvecbardown{\tau}-\cvec{c} - \cvec{g}$, and $\cvecbarup{d} = \cvecbarup{\tau} -\cvec{c} - \cvec{g}$.}
\begin{table}[tbp]
	\caption{Definitions of the variables in \bracketref{eq:genconstr} for velocity-, acceleration- and torque-based schemes.}
	\label{tab:unifiedControlProblem}
	\begin{tabular}{ M{0.03\textwidth}| M{0.12\textwidth}| M{0.12\textwidth}| M{0.12\textwidth}}
		&\textbf{velocity} &\textbf{acceleration} & \textbf{torque} \\ \hline 
		$\cmat{A}$ & $\cmat{J}_{eq}$ & $\cmat{J}_{eq}$ & $\cmat{J}_{eq} \cmat{M}^{-1}$ \\  \hline
		$\cvec{u}$ & $\cvecdot{q}$ & $\cvecddot{q}$ & $\cvec{\tau} - \cvec{c} - \cvec{g}$ \\ \hline 
		$\cvec{b}$ & $\cvec{r}_v$ & $\cvec{r}_a - \cmatdot{J}_{eq} \cvecdot{q}$ & $\cvec{r}_a - \cmatdot{J}_{eq} \cvecdot{q}$ \\ \hline
		$\cmat{C}$ & $\cmat{J}_{iq}$ & $\cmat{J}_{iq}$ & $\cmat{J}_{iq} \cmat{M}^{-1}$ \\ \hline
		$\cvecbardown{d}$ & $\cvecbardown{r}_v$ &  $\cvecbardown{r}_a - \cmatdot{J}_{iq} \cvecdot{q}$ & $\cvecbardown{r}_a - \cmatdot{J}_{iq} \cvecdot{q}$ \\ \hline
		$\cvecbarup{d}$ & $\cvecbarup{r}_v$ &  $\cvecbarup{r}_a - \cmatdot{J}_{iq} \cvecdot{q}$ & $\cvecbarup{r}_a - \cmatdot{J}_{iq} \cvecdot{q}$\\ \hline
	\end{tabular}
	\vspace{-10pt}
\end{table}
The problem of controlling a robot can therefore be reformulated as the problem of finding, for each instant of time, a suitable vector $\cvec{u}$ such that a certain number of constraints given in the form \bracketref{eq:genconstr} are satisfied. In case of redundant robots, infinite solutions exist to this problem. Thus, a solution can be found by solving a constrained optimization problem. In the context of redundancy resolution, the control effort is typically chosen as a cost, i.e.,
\begin{equation}
\label{eq:optProb}
	\begin{split}
	    &\min_{\cvec{u}} \ \frac{1}{2}\cvec{u}^{T} \cmat{H} \cvec{u} \\
	    \mathrm{s.t.} \; \cmat{A} &\cvec{u} = \cvec{b}, \quad \cvecbardown{d} \leq \cmat{C} \cvec{u} \leq \cvecbarup{d} ,
	\end{split}
\end{equation}
where $\cmat{H} \in \mathbb{R}^{n \times n}$ is an arbitrary, positive (semi-) definite weighting matrix. However, it is often desired to control the residual redundant DOFs of the robot by introducing a reference vector $\cvec{u}_r$ and solving the problem
\begin{equation}
\label{eq:optProb_u0}
	\begin{split}
	    \min_{\cvec{u}} & \ \frac{1}{2}(\cvec{u}-\cvec{u}_r)^{T} \cmat{H} (\cvec{u}-\cvec{u}_r)\\
	    \mathrm{s.t.} \; &\cmat{A} \cvec{u} = \cvec{b}, \quad \cvecbardown{d} \leq \cmat{C} \cvec{u} \leq \cvecbarup{d} .
	\end{split}
\end{equation}
This is of particular interest in acceleration- and torque-based schemes, where $\cvec{u}_r$ is typically modeled to damp internal motion of the manipulator \cite{Siciliano.1992}. Another common practice is to use $\cvec{u}_r$ to include an auxiliary optimization criteria (e.g., manipulability) via the GBP method \cite{liegeois1977automatic}. The choice of $\cmat{H}$ in \bracketref{eq:optProb} and \bracketref{eq:optProb_u0} is left to the user. However, specific metrics exist, which ensure dynamical consistency of the generated robot motion \cite{peters2008unifying, schettino2020geometrical}. Such metrics are used in Section \ref{sec:4_results_exp_iiwa}.

Omitting the inequality constraints, the solution for \bracketref{eq:optProb_u0} is given by \cite{liegeois1977automatic}
\begin{equation}
    \label{eq:optProb_singleTask}
	\cvec{u} = \cmat{A}^{\dagger H} \cvec{b} + \cmat{P}\cvec{u}_r,
\end{equation}
where $\cmat{A}^{\dagger H} = \cmat{H}^{-1} \cmat{A}^{T} \left(\cmat{A} \cmat{H}^{-1} \cmat{A}^{T}\right)^{-1}$ is the right pseudoinverse of $\cmat{A}$, weighted through $\cmat{H}$, and $\cmat{P} = \cmat{I} - \cmat{A}^{\dagger H}\cmat{A}$ is a projector in the null space of $\cmat{A}$. {\color{black} In \bracketref{eq:optProb_singleTask}, matrix $\cmat{A}$ is assumed to be full row-rank. Such hypothesis will be held throughout the paper also for matrix $\cmat{C}$. If this is not the case, singularity-robust techniques should be used in order to perform the pseudoinversion (see \cite{chiaverini97srtp, di2017comparison} for an overview).} 

{\color{black}
The optimization problem in \bracketref{eq:optProb} can be further extended to minimize 
\begin{equation}
\label{eq:optProb_multitask_min}
    \min_{\cvec{u}} \ \frac{1}{2}(\cvec{u} - \cvec{u}_r)^{T} \cmat{H} (\cvec{u} -\cvec{u}_r) \\
\end{equation}
while attempting to satisfying a set of multiple tasks (imposing equality and inequality constraints) specified with different levels of priorities
\begin{equation}
\label{eq:optProb_multitask}
	\begin{split}
    \mathrm{1^{st}\,level:} \; \cmat{A}_1 \cvec{u} &= \cvec{b}_1, \quad \cvecbardown{d}_1 \leq \cmat{C}_1 \cvec{u} \leq \cvecbarup{d}_1 \\
    &\vdots \quad , \qquad \quad \ \vdots \\
    \mathrm{k^{th}\,level:} \;\cmat{A}_k \cvec{u} &= \cvec{b}_k, \quad \cvecbardown{d}_k \leq \cmat{C}_k \cvec{u} \leq \cvecbarup{d}_k \\
    &\vdots \quad , \qquad \quad \ \vdots \\
    \mathrm{N^{th}\,level:} \; \cmat{A}_N \cvec{u} &= \cvec{b}_N, \quad \cvecbardown{d}_N \leq \cmat{C}_N \cvec{u} \leq \cvecbarup{d}_N, \\
	\end{split}
\end{equation}}
with $\cmat{A}_k$, $\cvec{b}_k$, $\cmat{C}_k$, $\cvecbardown{d}_k$ and $\cvecbarup{d}_k$ describing the equality and inequality constraints of the $k$th priority level (lower $k$ denoting higher priority). Omitting the inequality constraints, the solution to \bracketref{eq:optProb_multitask_min}-\bracketref{eq:optProb_multitask} can be computed recursively as \cite{Siciliano.1991}
\begin{equation}
\label{eq:sol_eq_k_multitask}
    \begin{cases}
        \cvec{u}_{0} = \cvec{0} \\
    	\cvec{u}_k = \cvec{u}_{k-1} + \left(\cmat{A}_k \cmat{P}_{k-1}\right)^{\dagger H} \left( \cvec{b}_k - \cmat{A}_k \cvec{u}_{k-1}\right) \\
    	\cvec{u} = \cvec{u}_N +{\color{black}\cmat{P}_N}\cvec{u}_r
    \end{cases},
\end{equation}
with $k = 1, ..., N$. The matrix $\cmat{P}_{k}  \in \mathbb{R}^{n \times n}$ in \bracketref{eq:sol_eq_k_multitask} is the well-known augmented projector from \cite{Siciliano.1991}, which can also be recursively computed as \cite{greville1960some}
\begin{equation}
\label{eq:proj_k_multitask}
    \begin{cases} 
        \cmat{P}_{0} = \cmat{I} \\
        \cmat{P}_{k} = (\cmat{I} - (\cmat{A}_k\cmat{P}_{k-1})^{\dagger H}\cmat{A}_k)\cmat{P}_{k-1}
\end{cases} .
\end{equation}
Finding a solution that fulfills also the inequality constraints is a more complex challenge, and it is the subject of the following section.

\section{Extended Saturation in the Null Space Method}
\label{sec:3_extendedSNS}
As remarked in Sect. \ref{sec:1intro}, the SNS algorithm \cite{flacco2015control} is a powerful tool for the (kinematic) control of redundant robots under hard joint constraints. In multiple iterations, it computes which joint requires the most scaling to be within its limits, and subsequently adds the inequality constraint of that limit to the equality constraints of the task. Moreover, a new scaling factor is computed for the task velocity reference, which would ensure all joints to be within their limits at the cost of a reduced speed in task execution. This cycle is repeated until either a solution that does not require task scaling is found, or all redundant DOFs of the robot are exhausted. In the latter case, the solution that requires the least scaling of the task (among the ones obtained in all the iterations) is returned. 

This section presents the proposed extension to the SNS algorithm, named \textit{extended SNS} (\textit{eSNS}), and its integration into the unified control framework introduced in Sect. \ref{sec:2c_genProb}. Thanks to this  framework, the eSNS algorithm can solve redundancy resolution problems defined at velocity, acceleration or torque level, indistinctly. Furthermore, compared to the original SNS, 
the proposed algorithm can handle arbitrary inequality constraints in the form \bracketref{eq:genconstr_ineq} on multiple priority levels. Finally, compared to pure joint velocity norm minimization, the eSNS attempts at minimizing the cost defined in \bracketref{eq:optProb_u0}, which includes the weighting matrix $\cmat{H}$ and the additional input vector $\cvec{u}_r$. 
\begin{table}[t!]
    \centering
	\caption{Definition of $\cvec{b}'$ and $\cvec{b}''$ in \bracketref{eq:sol_eq_k_multitask_eSNS} for velocity-, acceleration- and torque-based schemes}
	\label{tab:def_b_b}
    \begin{tabular}{ M{0.03\textwidth}| M{0.12\textwidth}| M{0.12\textwidth}| M{0.12\textwidth}}
		& velocity & acceleration & torque \\ \hline
		$\cvec{b}'$ & $\cvec{r}_v$ & $\cvec{r}_a$ & $\cvec{r}_a$ \\ \hline
		$\cvec{b}''$ & $\cvec{0}$ & $\cmatdot{J}_{eq} \cvecdot{q}$ & $\cmatdot{J}_{eq} \cvecdot{q}$ \\ \hline
    \end{tabular}
	\vspace{-10pt}
\end{table}

\subsection{Basic eSNS}
\label{sec:3_basicSNS}
\begin{algorithm}[t!]
	\caption{Basic Extended SNS algorithm for multiple priority levels}
	\label{alg:snsBasic}
	\begin{algorithmic}[1]
		\State $\cmat{P}_0 = \cmat{I}$, $\cvec{u}_0 = \cvec{0}$ 
		\For{$k = 1 \rightarrow N$}
		\State $\cmat{\tilde{A}}_k = \cmat{A}_k\cmat{P}_{k-1}$
		\State $\cmat{P}_k = \left(\cmat{I} - \cmat{\tilde{A}}_k^{\dagger H}\cmat{A}_k\right)\cmat{P}_{k-1}$
		\State $\cvec{u}_{sat} = \cvec{0}$
		\State $\cvec{u}' =  \cmat{\tilde{A}}_k^{\dagger H} \cvec{b}'_k$
		\State $\cvec{u}'' = \cmat{\tilde{A}}_k^{\dagger H} \left(-\cvec{b}''_k - \cmat{A}_k \cvec{u}_{k-1}\right)$
		\State $\cvec{u}''' = \cmat{P}_k \cvec{u}_{r,k}$
		\State $s^*_k = 0$
		\State $\cmat{C}^{sat}_{1 \rightarrow k} = \mathrm{null}$, $\cvec{d}^{sat}_{1 \rightarrow k} = \mathrm{null}$
		\Repeat
		\State limits\_violated = FALSE
		\State $\cvec{u}_k = \cvec{u}_{k-1} + \cvec{u}_{sat} + \cvec{u}' +  \cvec{u}'' + \cvec{u}'''$
		
        \State ${\color{black}\cvec{\rho}} = \cmat{C}_{1 \rightarrow k}\cvec{u}'$
		\State ${\color{black}\cvec{\phi}} = \cmat{C}_{1 \rightarrow k}\cvec{u}_k$
		\State ${\color{black}\cvec{\sigma} = \cvec{\phi} - \cvec{\rho}}$
		\If{$\exists i \in \left[1:l_k\right]: \left({\color{black}\phi^{i}} < \underline{d}^i_{1 \rightarrow k}\right) \vee \left({\color{black}\phi^{i}} > \overline{d}_{1 \rightarrow k}\right)$}
		\State limits\_violated = TRUE
		\State getTaskScalingFactor({\color{black}$\cvec{\rho}$,$\cvec{\sigma}$}) \Comment call alg. \ref{alg:taskScalingFactor}
		\If{$\mathrm{task \; scaling \; factor} > s^*_k$}
		\State $s^*_k = \mathrm{task \; scaling \; factor}$
		\State $\cvec{u}_{sat}^{*} = \cvec{u}_{sat}$
		\State $\cvec{u}'^* = \cvec{u}'$
		\State $\cvec{u}''^* = \cvec{u}''$
		\State $\cvec{u}'''^* = \cvec{u}'''$ 
		\EndIf
		\State $j = \mathrm{most \; critical \; constraint}$
		\State $\cmat{C}^{sat}_{1 \rightarrow k} \leftarrow \text{concatenate}(\cmat{C}^{sat}_{1 \rightarrow k}, \cvec{c}^{j}_{1 \rightarrow k})$
		\State $\cmat{\hat{P}}_{k-1} = \left(\cmat{I} - \left(\cmat{C}^{sat}_{1\rightarrow k}\cmat{P}_{k-1}\right)^{\dagger H}\cmat{C}^{sat}_{1\rightarrow k}\right)\cmat{P}_{k-1}$
		\State $\cmat{\hat{A}}_k = \cmat{A}_k \cmat{\hat{P}}_{k-1}$
		
		\If{$\mathrm{rank} (\cmat{\hat{A}}_k) \geq {\color{black}m_k}$}
		    \State $d^{j}_{1 \rightarrow k}  = \begin{cases}
		    \underline{d}^j_{1\rightarrow k}, &  {\color{black}\phi^j} < \underline{d}^j_{1\rightarrow k} \\
            \overline{d}^j_{1\rightarrow k}, & {\color{black}\phi^j} > \overline{d}^j_{1\rightarrow k}
		    \end{cases}$
		    \State $\cvec{d}^{sat}_{1 \rightarrow k} \leftarrow \text{concatenate}(\cvec{d}^{sat}_{1 \rightarrow k}, d^{j}_{1 \rightarrow k})$
		    \State $\cmat{\hat{C}}_k = \cmat{C}^{sat}_{1\rightarrow k}\cmat{P}_{k-1}$
		    \State $\cmat{\hat{P}}_k = \left(\cmat{I} - \cmat{\hat{A}}_k^{\dagger H}\cmat{A}_{k}\right)\cmat{\hat{P}}_{k-1}$
		    \State $\cvec{u}_{sat} = \cmat{\hat{C}}_k^{\dagger H} \left( \cvec{d}^{sat}_{1\rightarrow k} - \cmat{C}^{sat}_{1\rightarrow k} \cvec{u}_{k-1} \right)$
		    \State $\cvec{u}' = \cmat{\hat{A}}_k^{\dagger H} \cvec{b}'_k$
		    \State $\cvec{u}'' = \cmat{\hat{A}}_k^{\dagger H} \left(-\cvec{b}''_k - \cmat{A}_k \left(\cvec{u}_{k-1} + \cvec{u}_{sat} \right)\right)$
		    \State $\cvec{u}''' = \cmat{\hat{P}}_k \cvec{u}_{r,k}$
		\Else
		    \State $\cvec{u}_k = \cvec{u}_{k-1} + \cvec{u}^*_{sat} + s^*_k \cvec{u}'^* + \cvec{u}''^* + \cvec{u}'''^*$
		    \State limits\_violated = FALSE
		\EndIf
		\EndIf
		\Until limits\_violated = FALSE
		\EndFor
		\State $\cvec{u} = \cvec{u}_{N}$
	\end{algorithmic}
\end{algorithm}
With reference to Algorithm \ref{alg:snsBasic}, this section shows the steps of the eSNS in its basic version. 

{\color{black}First, the term $\cvec{b}_k$ from \bracketref{eq:sol_eq_k_multitask} can be decomposed as $\cvec{b}_k = \cvec{b}'_k-\cvec{b}''_k$. The definitions for $\cvec{b}'_k,\cvec{b}''_k\in \mathbb{R}^n$ can be found in Tab. \ref{tab:def_b_b} (where dependency on $k$ is omitted for brevity). Thus, the solution \bracketref{eq:sol_eq_k_multitask} for the $k$th level of priority considering only the equality constraints can be rewritten as}
\begin{equation}
    \label{eq:sol_eq_k_multitask_eSNS}
	\cvec{u}_k = \cvec{u}_{k-1} + \left(\cmat{A}_k \cmat{P}_{k-1}\right)^{\dagger H} \left(\cvec{b}'_k - \cvec{b}''_k - \cmat{A}_k \cvec{u}_{k-1}\right) + \cmat{P}_k\cvec{u}_{r,k}
\end{equation}
{\color{black}with} 
\begin{equation*}
    \cvec{u}_{r,k}=
    \begin{cases}
      \cvec{u}_r, & \text{if}\ k=N \\
      \cvec{0}, & \text{otherwise}.
    \end{cases}
\end{equation*}
To additionally ensure satisfaction of inequality constraints, the solution \bracketref{eq:sol_eq_k_multitask_eSNS} is expanded to the form (line 13)
\begin{equation}
    \label{eq:sol_k_multitask_eSNS}
	\cvec{u}_k = \cvec{u}_{k-1} + \cvec{u}_{sat} + \cvec{u}' + \cvec{u}'' + \cvec{u}''',
\end{equation}
with the following initialization (lines 5--8):
\begin{equation*}
    \begin{split}
	    \cvec{u}_{sat} &= \cvec{0} \\ 
	    \cvec{u}' &= \left(\cmat{A}_k \cmat{P}_{k-1}\right)^{\dagger H}\cvec{b}'_k \\
	    \cvec{u}'' &= \left(\cmat{A}_k \cmat{P}_{k-1}\right)^{\dagger H}\left(-\cvec{b}''_k - \cmat{A}_k (\cvec{u}_{k-1} + \cvec{u}_{sat})\right) \\
	    \cvec{u}''' &= \cmat{P}_k \cvec{u}_{r,k}.
	\end{split}
\end{equation*}
The vector $\cvec{u}_k$ is considered an admissible solution if it satisfies all the inequality constraints defined in the first $k$ levels of priority, namely
\begin{equation}
    \label{eq:first_k_ineq_constraints}
    \cvecbardown{d}_{1\rightarrow k} \leq \cmat{C}_{1\rightarrow k} \cvec{u}_k \leq \cvecbarup{d}_{1\rightarrow k},
\end{equation}
with $\cvecbardown{d}_{1\rightarrow k} = \begin{bmatrix} \cvecbardown{d}^{T}_1 & ... & \cvecbardown{d}^{T}_k \end{bmatrix}^{T}\in \mathbb{R}^{l_k}$, $\cvecbarup{d}_{1\rightarrow k} = \begin{bmatrix} \cvecbarup{d}^{T}_1 & ... & \cvecbarup{d}^{T}_k \end{bmatrix}^{T}\in \mathbb{R}^{l_k}$ and $\cmat{C}_{1\rightarrow k} = \begin{bmatrix} \cmat{C}^{T}_1 & ... & \cmat{C}^{T}_k \end{bmatrix}^{T} \in \mathbb{R}^{l_k \times n}$. If a solution $\cvec{u}_k$ is admissible (line 17), the algorithm moves to the next priority level. If, instead, some of the inequality constraints in \bracketref{eq:first_k_ineq_constraints} are violated, the eSNS starts/keeps iterating (line 11) to find a valid solution, eventually scaling the task velocity/acceleration in $\cvec{b}'_k$ by a factor $s_k \in [0,1]$ (line 41), if needed. At each iteration, the task scaling factor $s_k$ is computed using Algorithm \ref{alg:taskScalingFactor} ({\color{black}line 19}).
\begin{algorithm}[t!]
	\caption{Task scaling factor and most critical constraint computation}
	\label{alg:taskScalingFactor}
	\begin{algorithmic}[1]
		\Function{getTaskScalingFactor}{{\color{black}$\cvec{\rho}$,$\cvec{\sigma}$}}
		\For{$i = 1 \rightarrow l_k$}
		\State $\underline{s}^i = \left(\underline{d}_{1 \rightarrow k}^i - {\color{black}\sigma^i}\right) / {\color{black}\rho^i}$
		\State $\overline{s}^i = \left(\overline{d}_{1 \rightarrow k}^i - {\color{black}\sigma^i}\right) / {\color{black}\rho^i}$
		\If{$\underline{s}^i > \overline{s}^i$}
		\State $\mathrm{switch} (\underline{s}^i, \overline{s}^i)$
		\EndIf
		\EndFor
		\State $s_{\mathrm{max}} = \min_i \left\{ \cvecbarup{s} \right\}$
		\State $s_{\mathrm{min}} = \max_i \left\{ \cvecbardown{s} \right\}$
		\State $\mathrm{most \; critical \; constraint} = \mathrm{argmin}_i\left\{\cvecbarup{s}\right\}$
		\If{$s_{\mathrm{min}} > s_{\mathrm{max}} \vee s_{\mathrm{max}} < 0 \vee s_{\mathrm{min}} > 1$}
		\State $\mathrm{task \; scaling \; factor} = 0$
		\Else
		\State $\mathrm{task \; scaling \; factor} = \min \left\{ s_{\mathrm{max}}, 1 \right\}$
		\EndIf
		\EndFunction
	\end{algorithmic}
\end{algorithm}
The input arguments {\color{black} $\cvec{\rho}$ and $\cvec{\sigma}$} are given as (lines 14--16):
\begin{equation}
	\label{eq:alpha_beta}
	\begin{split}
		{\color{black}\cvec{\rho}} &= \cmat{C}_{1 \rightarrow k}\cvec{u}' \\
		{\color{black}\cvec{\sigma}} &= \cmat{C}_{1 \rightarrow k}\cvec{u}_k - {\color{black}\cvec{\rho}}.
	\end{split}
\end{equation} 
Although a new scaling factor is computed at every eSNS iteration, it is only applied when all redundant DOFs of the robot have been exhausted without finding a valid solution (line 41). In all other cases, it is only checked whether the current solution allows larger task scaling. In such a case, the solution (and the corresponding task scaling factor) is stored as the best found (lines 21--25). 

Algorithm \ref{alg:taskScalingFactor} also returns the most critical constraint, i.e., the violated inequality constraint that requires the smallest scaling factor $s_k$ to be satisfied. This constraint is used in the next phase of the algorithm, which is the saturation phase (lines 27--39). First, the most critical constraint is converted into an equality constraint
\begin{equation*}
    \label{eq:sat_eq_constr}
    (\cvec{c}^j_{1\rightarrow k})^T \cvec{u}_k = d^j_{1\rightarrow k},
\end{equation*}
with $j$ being the index of the most critical constraint and
\begin{equation*}
    d^j_{1\rightarrow k}=
    \begin{cases}
      \underline{d}^j_{1\rightarrow k}, & \text{if}\ (\cvec{c}^j_{1\rightarrow k})^T \cvec{u}_k < \underline{d}^j_{1\rightarrow k} \\
      \overline{d}^j_{1\rightarrow k}, & \text{if}\ (\cvec{c}^j_{1\rightarrow k})^T \cvec{u}_k > \overline{d}^j_{1\rightarrow k}
    \end{cases}.
\end{equation*}
Then, it is added to the \textit{saturation set}, which is the set of all converted inequalities, forming a system of equations that can be expressed as
\begin{equation}
    \label{eq:saturation_set}
    \cvec{C}^{sat}_{1\rightarrow k} \cvec{u}_k = \cvec{d}^{sat}_{1\rightarrow k}.
\end{equation}
Finally, the saturation commands \bracketref{eq:saturation_set} are enforced in the solution \bracketref{eq:sol_k_multitask_eSNS}, updating its terms as follows (lines 36--39):
\begin{equation}
    \label{eq:sol_k_multitask_eSNS_pieces}
    \begin{split}
	    \cvec{u}_{sat} &= \left( \cmat{C}^{sat}_{1\rightarrow k}\cmat{P}_{k-1}\right)^{\dagger H} \left( \cvec{d}^{sat}_{1\rightarrow k} - \cmat{C}^{sat}_{1\rightarrow k} \cvec{u}_{k-1} \right) \\ 
	    \cvec{u}' &= \left(\cmat{A}_k \cmat{\hat{P}}_{k-1}\right)^{\dagger H}\cvec{b}'_k \\
	    \cvec{u}'' &= \left(\cmat{A}_k {\cmat{\hat{P}}}_{k-1}\right)^{\dagger H}\left(-\cvec{b}''_k - \cmat{A}_k (\cvec{u}_{k-1} + \cvec{u}_{sat})\right) \\
	    \cvec{u}''' &= \cmat{\hat{P}}_k \cvec{u}_{r,k},
	\end{split}
\end{equation}
with
\begin{equation*}
    \begin{split}
        \cmat{\hat{P}}_{k-1} &= \left(\cmat{I} - \left(\cmat{C}^{sat}_{1\rightarrow k}\cmat{P}_{k-1}\right)^{\dagger H}\cmat{C}^{sat}_{1\rightarrow k}\right)\cmat{P}_{k-1} \\
	    \cmat{\hat{P}}_{k} &= \left(\cmat{I} - \left(\cmat{A}_{k}\cmat{\hat{P}}_{k-1}\right)^{\dagger H}\cmat{A}_{k}\right)\cmat{\hat{P}}_{k-1}
    \end{split}.
\end{equation*}
It should be noticed that enforcing a new saturation only makes sense if there are enough redundant DOFs left in the system to handle it. From an algebraic point of view, this condition is satisfied if $\text{rank} (\cmat{A}_k \cmat{\hat{P}}_{k-1}) \geq m_k$ (line 31), with $\cmat{A}_k \in \mathbb{R}^{m_k \times n}$  ($m_k \leq n$). 

\subsection{Relation to the original SNS algorithm}
\label{sec:3_relToOrig}
Analysing Algorithm \ref{alg:snsBasic}, a similar structure as the original SNS algorithm \cite{flacco2015control} can be recognized: after computing a first guess that satisfies the equality constraints, the solution is updated as long as one or more limits are found to be violated. Insertions to the saturation set happen using a policy based on the identification of the most critical constraint. Finally, task scaling is applied when all redundant DOFs have been used without finding a valid solution. It can  also be noticed from \bracketref{eq:sol_k_multitask_eSNS_pieces} that, as in the original SNS algorithm, an intermediate level of priority is imposed between the saturation commands $\cvec{u}_{sat}$ and the other terms introduced in \bracketref{eq:sol_k_multitask_eSNS}. This is guaranteed by the null space projectors  $\cmat{\hat{P}}_{k-1}$ and $ \cmat{\hat{P}}_{k}$. 

On the other hand, thanks to the special structure of $\cmat{\hat{C}}_k$ and $\cmat{\hat{P}}_{k-1}$, Algorithm \ref{alg:snsBasic} can perform saturation in any task space and, thus, handle arbitrary inequality constraints in the form \bracketref{eq:genconstr_ineq}. This is a key extension compared to the work in \cite{flacco2015control}, where only joint space inequality constraints have been handled. Moreover, the general cost \bracketref{eq:optProb_u0} is considered in Algorithm \ref{alg:snsBasic} through the use of general weighted pseudoinverses and the inclusion of the $\cvec{u}'''$ in \bracketref{eq:sol_k_multitask_eSNS_pieces}. In fact, specializing Algorithm \ref{alg:snsBasic} for velocity-based redundancy resolution, setting $\cmat{H} = \cmat{I} \in \mathbb{R}^{n \times n}$, $\cvec{u}_r=\cvec{0}$ and considering only joint space constraint with the highest priority ($\cmat{C}_{1 \rightarrow k} = \cmat{I} \in \mathbb{R}^{n \times n} \, \forall k=1,\dots, N$) would return the multi-task SNS algorithm in \cite{flacco2015control}.

To further highlight the relation between eSNS and SNS, the saturation commands $\cvec{u}_{sat}$ in \bracketref{eq:sol_k_multitask_eSNS_pieces} can be rewritten in the form
\begin{equation}
    \label{eq:sat_commands_snslike}
    \cvec{u}_{sat} = \left( \left( \cmat{I} - \cmat{W}_k \right)\cmat{C}_{1\rightarrow k}\cmat{P}_{k-1} \right)^{\dagger H} \cvec{d}_{sat},
\end{equation}
where
\begin{equation*}
    {d}^j_{sat} = 
    \begin{cases}
      \underline{d}^j_{1\rightarrow k} - (\cvec{c}^j_{1\rightarrow k})^T\cvec{u}_{k-1}, & \text{if}\ (\cvec{c}^j_{1\rightarrow k})^T \cvec{u}_k < \underline{d}^j_{1\rightarrow k} \\
      \overline{d}^j_{1\rightarrow k} - (\cvec{c}^j_{1\rightarrow k})^T\cvec{u}_{k-1}, & \text{if}\ (\cvec{c}^j_{1\rightarrow k})^T \cvec{u}_k > \overline{d}^j_{1\rightarrow k}\\ 
      0, & \text{otherwise}
    \end{cases},
\end{equation*}
from which it is easier to recognize the similarity with the joint velocity saturation in \cite{flacco2015control}. The diagonal matrix $\cmat{W}_k \in \mathbb{R}^{l_k \times l_k}$, whose diagonal has $0$ elements in the entries corresponding to the constraints in the saturation set and $1$ in all other entries, operates as a constraint selection matrix in \bracketref{eq:sat_commands_snslike} as in \cite{flacco2015control}. Also the projector $\cmat{\hat{P}}_{k-1}$ can be brought in a form that recalls the one in \cite{flacco2015control}
\begin{equation}
    \label{eq:nsproj_snslike}
    \cmat{\hat{P}}_{k-1} = \left(\cmat{I} - \left( \left( \cmat{I}-\cmat{W}_k \right) \cmat{C}_{1\rightarrow k}\cmat{P}_{k-1}\right)^{\dagger H}\cmat{C}_{1\rightarrow k} \right)\cmat{P}_{k-1}. \\
\end{equation}

\subsection{Fast-eSNS}
\label{sec:3_fasteSNS}

This section analyses the numerical performance of Algorithm \ref{alg:snsBasic}. The goal is to obtain a more efficient algorithm, which will be named \textit{Fast-eSNS}. The procedure will follow a similar reasoning as in \cite{flacco2013fast}. However, not all the machinery can be easily transferred to our general case, since \cite{flacco2013fast} exploits some specific properties of the standard Moore-Penrose pseudoinverse.

At each level of priority, the first operation that is critical for the computation time is the pseudoinversion of the matrix $\cmat{\tilde{A}}_k = \cmat{A}_k\cmat{P}_{k-1}$ (line 3 of Algorithm \ref{alg:snsBasic}). Different numerical methods exist for  computing the pseudoinverse of a matrix and the programmer must decide on a trade-off between speed and robustness. A common choice is to resort to the Singular Value Decomposition (SVD), which enables the analysis of singularities and the implementation of damped pseudoinversion methods. However, a faster method based on QR-Decomposition exists, as extensively pointed out in \cite{flacco2013fast, ziese2020redundancy}. Consider the QR-Decomposition of the product $\cmat{\tilde{A}}_k\cmat{H}^{-1}\cmat{\tilde{A}}_k^T$
\begin{equation*}
    \cmat{\tilde{A}}_k\cmat{H}^{-1}\cmat{\tilde{A}}_k^T = \cmat{Q}_k
    \begin{bmatrix}
        \cmat{R}_k \\ \cmat{0} 
    \end{bmatrix} =
    \begin{bmatrix}
        \cmat{Y}_k & \cmat{Z}_k
    \end{bmatrix}
    \begin{bmatrix}
    \cmat{R}_k \\ \cmat{0}
    \end{bmatrix},
\end{equation*}
with $\cmat{Y}_k \in \mathbb{R}^{n \times m_k}$ and $\cmat{Z_k}\in \mathbb{R}^{n \times (n-m_k)}$ orthogonal matrices composing $\cmat{Q}_k$, and $\cmat{R}_k \in \mathbb{R}^{m_k \times m_k}$ being upper triangular. Then, the weighted pseudoinverse of $\cmat{\tilde{A}}_k$ can be computed as
\begin{equation*}
    \cmat{\tilde{A}}_k^{\dagger H} = \cmat{H}^{-1}\cmat{\tilde{A}}^T_k \cmat{Y}_k\cmat{R}^{-T}_k,
\end{equation*}
which requires the inversion of an $m_k \times m_k$ upper triangular matrix only.

Proceeding with the analysis of Algorithm \ref{alg:snsBasic}, it can be easily recognized that the second expensive operation is the update of the solution $\cvec{u}_k$ (lines 36--39). This requires, at every iteration, the computation of the pseudoinverse of $\cmat{\hat{C}}_k = \cmat{C}^{sat}_{1 \rightarrow k}\cmat{P}_{k-1}$ and $\cmat{\hat{A}}_k = \cmat{A}_k\cmat{\hat{P}}_{k-1}$. However, the new solution can be computed  from the previous one by a rank-one update of the pseudoinverse computations. The first step is the rewriting of the general solution $\cvec{u}_k$ by using task augmentation \cite{chiacchio1991closed} as
\begin{align}
\label{eq:sol_k_task_augmentation}
	\begin{split}
	\cvec{u}_k &= \cvec{u}_{k-1} + \left( \cmat{A}_{TA,k} \cmat{P}_{k-1} \right)^{\dagger H}\cvec{b}_{TA,k} \\ 
	&= \cvec{u}_{k-1} +
	\begin{bmatrix}
	    \begin{pmatrix}
	        \cmat{A}_k \\ \cmat{C}^{sat}_{1 \rightarrow k} 
	    \end{pmatrix}
	    \cmat{P}_{k-1}  
	\end{bmatrix}^{\dagger H} 
	\begin{bmatrix} 
	    \cvec{b}'_{k} - \cvec{b}''_{k} - \cmat{A}_k \cvec{u}_{k-1} \\ \cvec{d}^{sat}_{1 \rightarrow k} - \cmat{C}^{sat}_{1 \rightarrow k} \cvec{u}_{k-1}
	\end{bmatrix}.
	\end{split}
\end{align}
It is possible to show that the result of \bracketref{eq:sol_k_task_augmentation} coincides with the one obtained from \bracketref{eq:sol_k_multitask_eSNS_pieces}, only if a feasible solution exists. However, the case of infeasible solutions is already excluded in Algorithm \ref{alg:snsBasic} by the condition  $\text{rank}(\cmat{A}_k \cmat{\hat{P}}_{k-1}) \geq m_k$. At each eSNS iteration, only one saturation command is imposed, corresponding to the most critical constraint $j$. Thus, \bracketref{eq:sol_k_task_augmentation} can be rewritten as
\begin{align*}
    \cvec{u}_k &= \cvec{u}_{k-1} \\
    &+
    \begin{bmatrix}
	    \begin{pmatrix}
	        \cmat{A}_{TA,k} \\ (\cvec{c}^{j}_{1 \rightarrow k})^T 
	    \end{pmatrix}
	    \cmat{P}_{k-1}  
	\end{bmatrix}^{\dagger H} 
	\begin{bmatrix} 
	    \cvec{b}'_{k} - \cvec{b}''_{k} - \cmat{A}_k \cvec{u}_{k-1} \\ \cvec{d}^{sat}_{1 \rightarrow k} - \cmat{C}^{sat}_{1 \rightarrow k} \cvec{u}_{k-1} \\ d^{j}_{1 \rightarrow k} - (\cvec{c}^{j}_{1 \rightarrow k})^T \cvec{u}_{k-1}
	\end{bmatrix},
\end{align*}
with $ \cmat{A}_{TA,k}$ initialized to $ \cmat{A}_k$ at the first iteration and redefined at the end of every cycle as:
\begin{equation}
    \label{eq:update_Ata}
    \cmat{A}_{TA,k} \leftarrow 
    \begin{bmatrix}
        \cmat{A}_{TA,k} \\ (\cvec{c}^{j}_{1 \rightarrow k})^T 
    \end{bmatrix}.
\end{equation}
It can be easily noticed that, at each new iteration, the matrix to pseudoinvert differs from the one at the previous iteration only by the appended row $(\cvec{c}^{j}_{1 \rightarrow k})^T\cmat{P}_{k-1}$. Therefore, a rank-one update strategy could be used. In particular, an update method for arbitrary weighting matrix has been derived from the algorithm in \cite{guo1986recursive} for appending a column. This yields
\begin{equation}
    \begin{split}
        \cvec{u}_k &= \cvec{u}_{k-1} + 
        \begin{bmatrix}
            \left(\cmat{I} - \cvec{\rchi} (\cvec{c}^j_{1 \rightarrow k})^T\right) \left(\cmat{A}_{TA,k}\cmat{P}_{k-1}\right)^{\dagger H} & \ \cvec{\rchi} 
        \end{bmatrix} \\
	    &\times
	    \begin{bmatrix}
	        \cvec{b}'_{k} - \cvec{b}''_{k} - \cmat{A}_k \cvec{u}_{k-1} \\ \cvec{d}^{sat}_{1 \rightarrow k} - \cmat{C}^{sat}_{1 \rightarrow k} \cvec{u}_{k-1} \\ d^{j}_{1 \rightarrow k} - (\cvec{c}^{j}_{1 \rightarrow k})^T \cvec{u}_{k-1}
	    \end{bmatrix},
	\end{split}
\end{equation}
with $ \cvec{\rchi} = \left((\cvec{c}^j_{1\rightarrow k})^T \cmat{\hat{P}}_{k}\right)^{\dagger H} \in \mathbb{R}^n$ defined as \textit{update vector}. At each new iteration the solution can then be updated as:
\begin{equation}
    \label{eq:sol_k_fast_eSNS_pieces}
    \begin{split}
        \cmat{\hat{P}}_{k} &\leftarrow \left(\cmat{I} - \cvec{\rchi} (\cvec{c}^{j}_{1 \rightarrow k})^T\right)\cmat{\hat{P}}_{k} \\
	    \cvec{u}_{sat} &\leftarrow \cvec{u}_{sat} + \cvec{\rchi} \left(d^j_{1\rightarrow k} - (\cvec{c}^{j}_{1 \rightarrow k})^T \left( \cvec{u}_{k-1} + \cvec{u}_{sat} \right) \right) \\ 
	    \cvec{u}' &\leftarrow \cvec{u}' - \cvec{\rchi}(\cvec{c}^{j}_{1 \rightarrow k})^T\cvec{u}' \\
	    \cvec{u}'' &\leftarrow \cvec{u}'' - \cvec{\rchi} (\cvec{c}^{j}_{1 \rightarrow k})^T\cvec{u}'' \\
	    \cvec{u}''' &\leftarrow \cmat{\hat{P}}_k \cvec{u}_{r,k}.
	\end{split}
\end{equation}
It should be noticed that only the weighted pseudoinverse of a vector is required to compute $\cvec{\rchi}$ and, thus, to update the solution according to \bracketref{eq:sol_k_fast_eSNS_pieces}. This allows significantly speeding up the update of the eSNS solution.

\subsection{Optimality of the eSNS solution}
\label{sec:3_optimality_sol}

This section focuses on the optimality properties of the eSNS. The analysis will refer to the algorithm introduced in Sect. \ref{sec:3_basicSNS}, but the same considerations would apply for the Fast-eSNS. With reference to the optimization problem \bracketref{eq:optProb_multitask}, it is easy to see that Algorithm \ref{alg:snsBasic} provides an optimal solution in case no saturation commands are required. In fact, the entire algorithm reduces to \bracketref{eq:sol_eq_k_multitask} in this trivial case. In case some inequality constraints are violated, however, the eSNS can enforce saturation commands at each level of priority following a specific policy, which is based on the iterative identification of the most critical constraint. Moreover, if at the $k$th level of priority it was not possible to find a feasible solution $\cvec{u_k}$, the eSNS applies a task scaling strategy. It should be noticed that the final scaling factor will also depend on the specific sequence with which the constraints were added to the saturation set. Thus, it can be concluded that, in general, there is no guarantee that the final saturation set and task scaling factor will produce an optimal solution. In other words, other admissible solutions might exist that return a smaller value of the cost function in \bracketref{eq:optProb_multitask} and/or higher scale factors.

To check whether a solution is optimal in some sense, a suitable optimization criterion must be defined. To this purpose, the following optimization problem is introduced to compute a solution at the $k$th level of priority
\begin{equation}
    \label{eq:opt_prob_k_Mscale}
	\begin{split}
	    \min_{\cvec{u}, s_k} \ \frac{1}{2}&(\cvec{u} - \cvec{u}_{r,k})^{T} \cmat{H} (\cvec{u} -\cvec{u}_{r,k}) + \frac{1}{2} M(1-s_k)^2\\
	    \mathrm{s.t.} \; &\cmat{A}_k \cvec{u} = s_k\cvec{b}'_k - \cvec{b}''_k \\ &\cmat{A}_{1 \rightarrow k-1} \cvec{u} = \cvec{b}'^{*}_{1 \rightarrow k-1} - \cvec{b}''_{1 \rightarrow k-1} \\ 
	    &\cvecbardown{d}_{1 \rightarrow k} \leq \cmat{C}_{1 \rightarrow k} \cvec{u} \leq \cvecbarup{d}_{1 \rightarrow k}\\
	    & 0 \leq s_k \leq 1
	\end{split},
\end{equation}
where $\cvec{b}'^{*}_{1 \rightarrow k-1} = \begin{bmatrix} s_1\cvec{b}'^{T}_1 & ... &  s_{k-1}\cvec{b}'^{T}_{k-1} \end{bmatrix}^{T}$ and  $M\gg1$ is a new (scalar) parameter used to weight the maximization of the scale factor $s_k$ in the range $[0, 1]$ with respect to the minimization of $(\cvec{u} - \cvec{u}_{r,k})^{T} \cmat{H} (\cvec{u} -\cvec{u}_{r,k})$. The problem \bracketref{eq:opt_prob_k_Mscale} can be rewritten as a standard quadratic problem with linear (equality and inequality) constraints
\begin{equation}
    \label{eq:opt_prob_k_xi}
    \begin{split}
	    &\min_{\cvec{\xi}} \ \frac{1}{2}\cvec{\xi}^{T} \cmat{\Theta} \cvec{\xi}\\
	    \mathrm{s.t.} \; &\cmat{\Lambda} \cvec{\xi} = \cvec{\beta}, \quad \cmat{\Gamma}\cvec{\xi} \leq \cvec{\delta},
	\end{split}
\end{equation}
with
\begin{align*}
    &\cvec{\xi} &&= 
    \begin{bmatrix}
        \cvec{u} - \cvec{u}_{r,k} \\ 1-s_k
    \end{bmatrix}, \cmat{\Theta} =
    \begin{bmatrix}
        \cmat{H} & \cvec{0} \\ \cvec{0} & M
    \end{bmatrix}, \cmat{\Lambda} = 
    \begin{bmatrix}
        \cmat{A}_k & \cvec{b}'_k \\ \cmat{A}_{1 \rightarrow k-1} & \cvec{0}
    \end{bmatrix}, \\
    &\cvec{\beta} &&=
    \begin{bmatrix}
        \cvec{b}'_k - \cvec{b}''_k - \cmat{A}_k\cvec{u}_{r,k} \\ 
        \cvec{b}'^{*}_{1 \rightarrow k-1} - \cvec{b}''_{1 \rightarrow k-1} - \cmat{A}_{1 \rightarrow k-1}\cvec{u}_{r,k}
    \end{bmatrix}, \\
	&\cmat{\Gamma} &&=
	\begin{bmatrix}
	    -\cmat{C}_{1 \rightarrow k} & \cvec{0} \\ \cmat{C}_{1 \rightarrow k} & \cvec{0} \\ \cvec{0} & 1 \\ \cvec{0} & -1
	\end{bmatrix}, \quad \cvec{\delta} =
	\begin{bmatrix}
	    -\cvecbardown{d}_{1 \rightarrow k} + \cmat{C}_{1 \rightarrow k}\cvec{u}_{r,k} \\ \cvecbarup{d}_{1 \rightarrow k} - \cmat{C}_{1 \rightarrow k}\cvec{u}_{r,k} \\ 1 \\ 0
    \end{bmatrix}. 
\end{align*}
Necessary and sufficient optimality conditions for the problem \bracketref{eq:opt_prob_k_xi} are given by the well-known Karush-Kuhn-Tucker (KKT) criteria \cite{kuhn1951tucker}: 
\begin{subequations}
    \label{eq:kkt}
    \begin{align}
        \cmat{\Theta}\cvec{\xi} + \cmat{\Lambda}^{T}\cvec{\lambda} + \cmat{\Gamma}^{T}\cvec{\mu} &= \cvec{0}  \label{eq:kkt_a} \\
	    {\color{black}\cvec{\mu}^T} \left(\cmat{\Gamma}\cvec{\xi} - \cvec{\delta} \right) &= 0 \label{eq:kkt_b} \\
	    \cmat{\Lambda}\cvec{\xi} &= \cvec{\beta} \label{eq:kkt_c} \\
	    \cmat{\Gamma}\cvec{\xi} &\leq \cvec{\delta} \label{eq:kkt_d} \\
	    \cvec{\mu} &\geq \cvec{0}, \label{eq:kkt_e}
	\end{align}
\end{subequations}
where $\cvec{\lambda}\in \mathbb{R}^{m_1 + ... + m_k}$ and $\cvec{\mu}\in \mathbf{R}^{2l_k+2}$ are the Lagrange multipliers associated with the equality and inequality constraints, respectively.

In order to analyse the optimality properties of the eSNS, start considering condition \bracketref{eq:kkt_c}. This imposes the satisfaction of the $k$th equality constraint, possibly obtained through task scaling. Furthermore, it enforces that the $(k-1)$ higher priority equality constraints are satisfied, preserving the task scaling factors obtained by the eSNS up to the $(k-1)$th level of priority. Reviewing the structure of the eSNS, it can be easily stated that the satisfaction of the $k$th equality constraint is guaranteed at each iteration by construction of the algorithm. Moreover, the use of the projector $\cmat{\hat{P}}_{k-1}$ in the solution ensures that the satisfaction of higher priority equality constraints (each considered with its computed task scaling factor) is preserved. In the light of the considerations above, it is possible to extract from \bracketref{eq:kkt_a} an expression related only to the constraints of the $k$th priority level. This is achieved via left multiplication of \bracketref{eq:kkt_a} by the $(n+1) \times (n+1)$ matrix
\begin{equation*}
    \cmat{\Psi}_{k-1} = 
    \begin{bmatrix} 
        \cmat{P}^{T}_{k-1} & \cvec{0} \\ \cvec{0} & 1
    \end{bmatrix}.
\end{equation*}
Recalling that $\cmat{A}_{1 \rightarrow k-1}\cmat{P}_{k-1} = \cmat{0}$, the above-mentioned multiplication yields
\begin{equation}
    \label{eq:kkt_a_p}
    \cmat{\tilde{\Theta}}\cvec{\xi} + \cmat{\tilde{\Lambda}}^{T}\cvec{\tilde{\lambda}} + \cmat{\tilde{\Gamma}}^{T}\cvec{\mu} = \cvec{0},
\end{equation}
where $\cmat{\tilde{\Theta}} = \cmat{\Psi}_{k-1}\cmat{\Theta}$, $\cmat{\tilde{\Lambda}} = \begin{bmatrix} \cmat{A}_k\cmat{P}_{k-1} & \cvec{b}'_k \end{bmatrix}$, $\cmat{\tilde{\Gamma}} = \cmat{\Gamma}\cmat{\Psi}^{T}_{k-1}$, and $\cvec{\tilde{\lambda}} \in \mathbb{R}^{m_k}$ contains the first $m_k$ elements of $\cvec{\lambda}$.

Another condition that is automatically satisfied inside the eSNS is \bracketref{eq:kkt_d}. Indeed, task-related inequality constraints are checked at every eSNS iteration. Moreover, the constraints on $s_k$ are satisfied by construction of Algorithm \ref{alg:taskScalingFactor}, which already outputs task scaling factors in the range $[0,1]$. 

It can be noticed that, for the constraints belonging to the saturation set, also condition \bracketref{eq:kkt_b} is automatically satisfied inside the eSNS. Indeed, for a generic constraint $j$ in the set, it is $\cvec{\gamma}^j\cvec{\xi} - \delta^j = 0$, regardless of the value of $\mu^j \geq 0$ (from \bracketref{eq:kkt_e}). Satisfying \bracketref{eq:kkt_b} for those constraints that are not in the saturation set requires that the corresponding elements of $\cvec{\mu}$ are null. Defining the $(n+1) \times (n+1)$ matrix
\begin{equation*}
    \cmat{\hat{\Psi}}_{k-1} = 
    \begin{bmatrix} 
        \cmat{\hat{P}}^{T}_{k-1} & \cvec{0} \\ \cvec{0} & 1
    \end{bmatrix}
\end{equation*}
allows extracting from \bracketref{eq:kkt_a} an expression related only to constraints of the $k$th priority level that are not in the saturation set. This is achieved via left multiplication by $\cmat{\hat{\Psi}}_{k-1}$. Recalling that $\cmat{A}_{1 \rightarrow k-1}\cmat{\hat{P}}_{k-1} = \cmat{0}$ and that the components of $\cvec{\mu}$ associated to non-saturated constraints are null, the multiplication yields
\begin{equation}
    \label{eq:kkt_a_nonsat}
    \cmat{\hat{\Theta}}\cvec{\xi} + \cmat{\hat{\Lambda}}^{T} \cvec{\tilde{\lambda}} = \cvec{0},
\end{equation}
where $\cmat{\hat{\Theta}} = \cmat{\hat{\Psi}}_{k-1}\cmat{\Theta}$ and $\cmat{\hat{\Lambda}} = \begin{bmatrix} \cmat{A}_k\cmat{\hat{P}}_{k-1} & \cvec{b}'_k \end{bmatrix}$. Since the possible rank-deficiency of $\cmat{A}_k \cmat{\hat{P}}_{k-1}$ is already checked at each eSNS iteration, \bracketref{eq:kkt_a_nonsat} admits always a solution $\cvec{\tilde{\lambda}}$, which can be computed as
\begin{equation}
    \label{eq:kkt_lambda_sol}
    \cvec{\tilde{\lambda}} = - \left( \cmat{\hat{\Lambda}}^{\dagger\Theta} \right)^{T} \cmat{\hat{\Theta}}\cvec{\xi}.
\end{equation}
Note that the computation of $\cvec{\tilde{\lambda}}$ from \bracketref{eq:kkt_lambda_sol} is not expensive. Indeed, given the structure of $\cmat{\hat{\Lambda}}$, the matrix $\cmat{\hat{\Lambda}}^{\dagger\Theta}$ can be obtained as rank-one update of $(\cmat{A}_k\cmat{\hat{P}}_{k-1})^{\dagger\Theta}$, which is already computed at every eSNS iteration. Substituting \bracketref{eq:kkt_lambda_sol} in \bracketref{eq:kkt_a_p} yields
\begin{equation}
    \label{eq:kkt_gammamu}
    \cmat{\tilde{\Gamma}}^{T}\cvec{\mu} = -\left( \cmat{\tilde{\Theta}} - \cmat{\tilde{\Lambda}}^{T} \left( \cmat{\hat{\Lambda}}^{\dagger\Theta} \right)^{T} \cmat{\hat{\Theta}} \right) \cvec{\xi}.
\end{equation}
As already mentioned, condition \bracketref{eq:kkt_b} imposes that the elements of $\cvec{\mu}$ associated with the constraints that are not in the saturation set are null. Therefore, the vector $\cvec{\mu}$ can be partitioned as $\cvec{\mu} = \begin{bmatrix} {\cvec{\mu}^{sat}}^T & \cvec{0} \end{bmatrix}^{T} $, where $\cvec{\mu}^{sat} \in \mathbb{R}^{l_{sat}} $ collects the Lagrange multipliers associated with the saturated constraints. Thus, \bracketref{eq:kkt_gammamu} can be rewritten as
\begin{equation*}
    \label{eq:kkt_gammasatmusat}
    \left( \cmat{\tilde{\Gamma}}^{sat}\right)^{T}\cvec{\mu}^{sat} = -\left( \cmat{\tilde{\Theta}} - \cmat{\tilde{\Lambda}}^{T} \left( \cmat{\hat{\Lambda}}^{\dagger\Theta} \right)^{T} \cmat{\hat{\Theta}} \right) \cvec{\xi},
\end{equation*}
where $\cmat{\tilde{\Gamma}}^{sat} \in \mathbb{R}^{l_{sat} \times (n+1)}$ is the matrix obtained by extracting the rows of $\cmat{\tilde{\Gamma}}$ associated with the saturated constraints. 
{\color{black} Given the linear independence of the rows of $\cmat{C}_{1 \rightarrow k}$ (as assumed in Sect. \ref{sec:2c_genProb}), the matrix $\cmat{\tilde{\Gamma}}^{sat}$ can be considered as full row-rank. Thus, $\cvec{\mu}^{sat}$ can be computed as}
\begin{equation}
    \label{eq:kkt_musat}
\cvec{\mu}^{sat} = -\left(  \left( \cmat{\tilde{\Theta}}^{T} - \cmat{\hat{\Theta}}^T \cmat{\hat{\Lambda}}^{\dagger \Theta} \cmat{\tilde{\Lambda}} \right) \left( \cmat{\tilde{\Gamma}}^{sat} \right)^{\dagger \Theta} \right)^{T} \cvec{\xi}.
\end{equation}
Therefore, it can be concluded that the eSNS provides an optimal solution to the problem \bracketref{eq:opt_prob_k_Mscale} if and only if the components of $\cvec{\mu}^{sat}$ in \bracketref{eq:kkt_musat} are non-negative. Note that the evaluation of \bracketref{eq:kkt_musat} increases the computational burden of the algorithm. Indeed, an additional pseudoinverse operation is required for computing $(\cmat{\tilde{\Gamma}}^{sat})^{\dagger \Theta}$. Nevertheless, the optimality check can be significantly simplified by following similar reasoning as in \cite{flacco2013optimal}, i.e., by considering that the eSNS is already able to output a task scaling factor close to $1$ (its maximum). Thus, the optimality request on the scaling factor can be removed and the following simplified QP problem can be considered:
\begin{equation}
    \label{eq:opt_prob_reduced}
    	\begin{split}
	    \min_{\cvec{u}} \ \frac{1}{2}&(\cvec{u} - \cvec{u}_{r,k})^{T} \cmat{H} (\cvec{u} -\cvec{u}_{r,k}) \\
	    \mathrm{s.t.} \; &\cmat{A}_k \cvec{u} = s_k\cvec{b}'_k - \cvec{b}''_k \\ &\cmat{A}_{1 \rightarrow k-1} \cvec{u} = \cvec{b}'^{*}_{1 \rightarrow k-1} - \cvec{b}''_{1 \rightarrow k-1} \\ 
	    &\cvecbardown{d}_{1 \rightarrow k} \leq \cmat{C}_{1 \rightarrow k} \cvec{u} \leq \cvecbarup{d}_{1 \rightarrow k}
	\end{split},
\end{equation}
where $s_k$ is the scaling factor obtained from the eSNS. By setting 
\begin{align*}
    &\cvec{\xi} &&= \cvec{u} - \cvec{u}_{r,k}, \quad \cmat{\Theta} = \cmat{H}, \quad \cmat{\Lambda} = 
	\begin{bmatrix}
	    \cmat{A}_k \\ \cmat{A}_{1 \rightarrow k-1}
	 \end{bmatrix},\\
	 &\cvec{\beta} &&= 
	 \begin{bmatrix}
	    s_k\cvec{b}'_k - \cvec{b}''_k -\cmat{A}_k\cvec{u}_{r,k} \\ \cvec{b}'^{*}_{1 \rightarrow k-1} - \cvec{b}''_{1 \rightarrow k-1} - \cmat{A}_{1 \rightarrow k-1}\cvec{u}_{r,k}
	 \end{bmatrix}, \\
	 &\cmat{\Gamma} &&=
	 \begin{bmatrix}
	    -\cmat{C}_{1 \rightarrow k} \\ \cmat{C}_{1 \rightarrow k}
	 \end{bmatrix}, 
	 \cvec{\delta} = 
	 \begin{bmatrix}
	    -\cvecbardown{d}_{1 \rightarrow k} + \cmat{C}_{1 \rightarrow k}\cvec{u}_{r,k} \\ \cvecbarup{d}_{1 \rightarrow k} - \cmat{C}_{1 \rightarrow k}\cvec{u}_{r,k}
	 \end{bmatrix}, \quad 
\end{align*}
problem \bracketref{eq:opt_prob_reduced} can be brought to the standard form \bracketref{eq:opt_prob_k_xi} and the same analysis based on the KKT conditions can be conducted as in the previous case. Again, the only condition that needs to be checked is the non-negativeness of the components of the multipliers $\cvec{\mu}^{sat}$. These can be computed by following the same steps shown in the previous case. As a result, \bracketref{eq:kkt_musat} becomes
\begin{align*}
    \cvec{\mu}^{sat} = - \bigg( \cmat{H} \Big( \cmat{I} - \big( &\cmat{A}_k\cmat{\hat{P}}_{k-1} \big)^{\dagger H} \cmat{A}_k \Big) \\    
    &\times \big( \cmat{\Gamma}^{sat} \cmat{P}_{k-1} \big)^{\dagger H} \bigg)^{T} \left( \cvec{u} - \cvec{u}_{r,k} \right).
\end{align*}
Moreover, by considering the particular structure of $\cmat{\Gamma}$ and defining the auxiliary vector $\cvec{\tilde{\mu}}^{sat}$ as
\begin{equation}
    \label{eq:mu_tildesat}
    \begin{split}
        \cvec{\tilde{\mu}}^{sat} = - \bigg( \cmat{H} \Big( \cmat{I} - \big( &\cmat{A}_k\cmat{\hat{P}}_{k-1} \big)^{\dagger H}  \cmat{A}_k \Big) \\
        &\times \big( \cmat{C}^{sat}_{1 \rightarrow k} \cmat{P}_{k-1} \big)^{\dagger H} \bigg)^T \left( \cvec{u} - \cvec{u}_{r,k} \right),
    \end{split}
\end{equation}
it is possible to compute the components of $\cvec{\mu}^{sat}$ as
\begin{equation*}
    \mu^{sat,j} = 
    \begin{cases}
        -\tilde{\mu}^{sat,j}, & \text{if} \ (\cvec{c}^{sat,j}_{1\rightarrow k})^T \cvec{u}_k = \underline{d}^{sat,j}_{1\rightarrow k}\\
        ~~ \tilde{\mu}^{sat,j},  &\text{if}\ (\cvec{c}^{sat,j}_{1\rightarrow k})^T \cvec{u}_k = \overline{d}^{sat,j}_{1\rightarrow k}
    \end{cases},
\end{equation*}
with $j=1,\dots,l_{sat}$. Note that the evaluation of \bracketref{eq:mu_tildesat} requires very limited computational effort, since both $( \cmat{A}_k\cmat{\hat{P}}_{k-1})^{\dagger H}$ and $( \cmat{C}^{sat}_{1 \rightarrow k}\cmat{P}_{k-1})^{\dagger H}$ are already computed inside the eSNS. Analogously, using the considerations from Sect. \ref{sec:3_relToOrig}, it is also possible to define the auxiliary vector $\cvec{\tilde{\mu}}$ as
\begin{equation}
    \begin{split}
        \label{eq:mu_tilde}
        \cvec{\tilde{\mu}} =  - \bigg( &\cmat{H} \Big( \cmat{I} - \big( \cmat{A}_k \cmat{\hat{P}}_{k-1} \big)^{\dagger H}  \cmat{A}_k \Big) \\
        &\times  \Big( \big( \cmat{I}-\cmat{W}_k\big)\cmat{C}_{1 \rightarrow k} \cmat{P}_{k-1} \Big)^{\dagger H} \bigg)^T \left( \cvec{u} - \cvec{u}_{r,k} \right),
    \end{split}
\end{equation}
and then compute all the components of $\cvec{\mu}$ as follows:
\begin{equation}
    \label{eq:mu_from_mutilde}
    \begin{alignedat}{2}
        \mu^j & = \mu^{j+l_k} = \tilde{\mu}^{j} = 0,  &&\text{if} \ w^{j,j}_k = 1 \\
        \mu^j &= -\tilde{\mu}^j, \mu^{j+l_k} = 0, \ &&\text{if} \ w^{j,j}_k = 0 \ \text{AND} \ (\cvec{c}^j_{1\rightarrow k})^T \cvec{u}_k = \underline{d}^j_{1\rightarrow k}\\ 
        \mu^j &= 0, \mu^{j+l_k} = \tilde{\mu}^j,  &&\text{if} \ w^{j,j}_k = 0 \ \text{AND} \ (\cvec{c}^j_{1\rightarrow k})^T \cvec{u}_k = \overline{d}^j_{1\rightarrow k} \\
    \end{alignedat}
\end{equation}
with $j = 1, \dots, l_k$ and $w^{j,j}$ being the $j$th element of the diagonal of $\cmat{W}_k$. From \bracketref{eq:mu_tilde} and \bracketref{eq:mu_from_mutilde}, it is easier to see that the results of this section extend the work in \cite{flacco2013optimal, flacco2015control}. Indeed, in the special case in which $\cmat{H} = \cmat{C}_{1 \rightarrow k} = \cmat{I} \in \mathbb{R}^{n \times n}, \cvec{u}_{r,k} = \cvec{0} \ \forall k=1,\dots, N$, \bracketref{eq:mu_tilde} reduces to
\begin{equation*}
    \cvec{\tilde{\mu}} =  - \left( \left( \cmat{I} - \left( \cmat{A}_k \cmat{\hat{P}}_{k-1} \right)^{\dagger I}  \cmat{A}_k \right) \left( \left( \cmat{I}-\cmat{W}_k\right) \cmat{P}_{k-1} \right)^{\dagger I} \right)^T \cvec{u},
\end{equation*}
which, imposing $\cmat{A}_k = \cmat{J}_k$ (velocity-based redundancy resolution), returns the auxiliary vector introduced in \cite{flacco2013optimal}.

\subsection{Opt-eSNS}
\label{sec:3_opteSNS}
\begin{algorithm}[t!]
	\caption{Opt-eSNS algorithm for multiple priority levels}
	\label{alg:opteSNS}
	\begin{algorithmic}[1]
		\State $\cmat{P}_0 = \cmat{I}$, $\cvec{u}_0 = \cvec{0}$ 
		\For{$k = 1 \rightarrow N$}
		
		\State $\cmat{C}^{sat}_{1 \rightarrow k} \leftarrow \cmat{C}^{\{ sat \}^-_k}_{1 \rightarrow k}$
		\State $\cvec{d}^{sat}_{1 \rightarrow k} \leftarrow \cvec{d}^{\{ sat \}^-_k}_{1 \rightarrow k}$
		
		\State $\cmat{\hat{A}}_k = \cmat{A}_k \cmat{\hat{P}}_{k-1}$
		\State $\cmat{\hat{C}}_k = \cmat{C}^{sat}_{1\rightarrow k}\cmat{P}_{k-1}$
		\State $\cmat{P}_k = \left(\cmat{I} - \cmat{\tilde{A}}_k^{\dagger H}\cmat{A}_k\right)\cmat{P}_{k-1}$
		\State $\cmat{\hat{P}}_{k-1} = \left( \cmat{I} - \left( \cmat{C}^{sat}_{1\rightarrow k} \cmat{P}_{k-1} \right)^{\dagger H} \cmat{C}^{sat}_{1\rightarrow k} \right) \cmat{P}_{k-1}$
        \State $\cmat{\hat{P}}_k = \left( \cmat{I} - \left( \cmat{A}_{k}\cmat{\hat{P}}_{k-1} \right)^{\dagger H} \cmat{A}_{k} \right) \cmat{\hat{P}}_{k-1}$

		\State $\cvec{u}_{sat} = \cmat{\hat{C}}_k^{\dagger H} \left( \cvec{d}^{sat}_{1\rightarrow k} - \cmat{C}^{sat}_{1\rightarrow k} \cvec{u}_{k-1} \right)$
		\State $\cvec{u}' = \cmat{\hat{A}}_k^{\dagger H} \cvec{b}'_k$
	    \State $\cvec{u}'' = \cmat{\hat{A}}_k^{\dagger H} \left(-\cvec{b}''_k - \cmat{A}_k \left(\cvec{u}_{k-1} + \cvec{u}_{sat} \right)\right)$
		\State $\cvec{u}''' = \cmat{\hat{P}}_k \cvec{u}_{r,k}$
		
		\State $s^*_k = 0$

		\Repeat
		\State $... (\text{same code as Alg. \ref{alg:snsBasic}})$
		\State $\cvec{\tilde{\mu}}^{sat} = - \left( \cmat{H} \left( \cmat{I} - \cmat{\hat{A}}_k^{\dagger H} \cmat{A} \right) \cmat{\hat{C}}_k^{\dagger H} \right)^T \left( \cvec{u} - \cvec{u}_{r,k} \right)$
		
		\For{$j = 1 \rightarrow l_{sat}$}
			\If{$ \begin{pmatrix}
			 (\cvec{c}^{sat,j}_{1\rightarrow k})^T \cvec{u}_k = \underline{d}^{sat,j}_{1\rightarrow k} \text{ AND } \tilde{\mu}^{sat,j} > 0 \\ \text{ OR } \\ (\cvec{c}^{sat,j}_{1\rightarrow k})^T \cvec{u}_k = \overline{d}^{sat,j}_{1\rightarrow k} \text{ AND }\tilde{\mu}^{sat,j} < 0
			\end{pmatrix} $}
			\State
			\State $\{ sat \}_k \leftarrow \{ sat \}_k \backslash \{ j \}$
			\State limits\_violated = TRUE
		    \EndIf
		\EndFor
		\Until limits\_violated = FALSE
		\State $\{ sat \}^-_k = \{ sat \}_k$
		\EndFor
		\State $\cvec{u} = \cvec{u}_{N}$
	\end{algorithmic}
\end{algorithm}
Following the analysis from the previous section, it is possible to develop a new variant of the eSNS algorithm, named \textit{Opt-eSNS}, which can always guarantee optimality of the solution. The Opt-eSNS is derived from the basic one by simply adding the optimality check on $\cvec{\mu}^{sat}$ after every iteration (see Algorithm \ref{alg:opteSNS}). This allows constraints to also be removed from the saturation set, when the associated Lagrange multipliers are negative. In view of \bracketref{eq:mu_tildesat}, the optimality check can be performed directly on the components of $\cvec{\tilde{\mu}}^{sat}$. Furthermore, the possibility of removing constraints from the saturation set allows the initialization of $\cmat{C}^{sat}_{1 \rightarrow k}$ and $\cvec{d}^{sat}_{1 \rightarrow k}$ based on the result of the previous execution of the algorithm (\textit{warm start}). Considering that the commands $\cvec{b}_k$, $\cvecbardown{d}_{1 \rightarrow k}$, $\cvecbarup{d}_{1 \rightarrow k}$ are typically smooth, it is reasonable to assume that two consecutive solutions will have similar saturation sets at each level of priority. Thus, indicating with $\{ sat \}^-_k$ the set of indices associated with the saturated commands of the $k$th priority level at the previous execution of the algorithm, it is possible to initialize $\cmat{C}^{sat}_{1 \rightarrow k}$ and $\cvec{d}^{sat}_{1 \rightarrow k}$ by extracting from $\cmat{C}_{1 \rightarrow k}$, $\cvecbardown{d}_{1 \rightarrow k}$ and $\cvecbarup{d}_{1 \rightarrow k}$ the rows and the components according to the indices in $\{ sat \}^-_k$. This operation is carried out at lines 3--4 of Algorithm \ref{alg:opteSNS}. Then, the set $\{ sat \}^-_k$ is updated at line 26, after the solution for the $k$th priority level has been computed. This kind of initialization of the saturation set typically reduces the total number of iterations inside the algorithm, therefore resulting in a faster computation (see the results of Sect. \ref{sec:4_results_sim_diiwa}). Algorithm \ref{alg:opteSNS} iterates until an optimal solution is found. Therefore, a more suitable name for the boolean variable used at lines 22 and 25 would be \textit{non\_optimal\_solution}.

It is finally worth considering whether the adoption of the Opt-eSNS is preferable over the employment of state-of-the-art QP solvers, which can directly solve the problem \bracketref{eq:opt_prob_k_Mscale} for each level of priority. First, it should be noticed that the dimension of the set of constraints significantly increases as the dimension of $\cmat{A}_{1 \rightarrow k-1}$ increases in \bracketref{eq:opt_prob_k_Mscale} at each new priority level. This makes state-of-the-art QP solvers become slower as $k$ increases. On the other hand, the use of null space projectors in the Opt-eSNS limits the dimension of the matrices to be (pseudo-)inverted, typically resulting in faster computation times (see Sect. \ref{sec:4_results_sim_diiwa} for numeric results). Furthermore, although it is normally possible to find a good range of values, the choice of $M$ might result not to be trivial and dependent on the specific problem to solve. Values that are not big enough could lead to conservative scale factors. On the contrary, too large values can undermine the numerical stability of the solver and lead to infeasible problems (see Sect. \ref{sec:4_results_sim_iiwa}). Not being dependent on such parameter is then an additional benefit of the proposed Opt-eSNS algorithm.

\begin{figure*}[t!]
	\centering
	\includegraphics[width=0.6\textwidth]{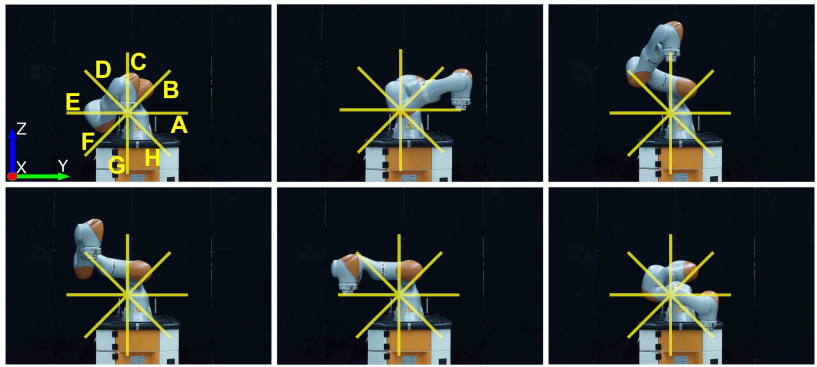}
    \caption{LBR iiwa moving on a star-like path during the experiments of Sect. \ref{sec:4_results_exp_iiwa}. The path is defined on the $YZ$ plane. Each star segment (indicated with a capital letter) has a length of $24$ cm.}
	\label{fig:iiwaMotion}
\end{figure*}
\section{Results}
\label{sec:4_results}
This section reports experiments and simulations to numerically support the effectiveness of the proposed framework in solving (prioritized) redundancy resolution problems defined at velocity, acceleration or torque level. The benefits of the variants presented in Sect. \ref{sec:3_fasteSNS} and Sect. \ref{sec:3_opteSNS} are also highlighted, compared to basic eSNS algorithm introduced in Sect. \ref{sec:3_basicSNS}. The presented results are obtained through experiments carried out on a KUKA LBR iiwa 7-DOFs robot (Fig. \ref{fig:iiwaMotion}), as well as through simulations involving a highly-redundant mobile dual-arm system (Fig. \ref{fig:experiment3diiwa_seq}).

\begin{table}[t!]
	\centering
	\caption{Initial configuration and joint limits for the LBR iiwa. The first joint is at the base of the robot.}
	\label{tab:jointPosLim}
	\begin{tabularx}{0.49\textwidth}{c|c|c|c|c}
		Joint & Position Lim. & Velocity Lim.& Acceler. Lim. & Initial Pos.\\ 
		nr.& $\cvecbardown{p},\cvecbarup{p}$ [rad] &  $\cvecbardown{v},\cvecbarup{v}$ [rad/s] & $\cvecbardown{a},\cvecbarup{a}$ [rad/s\textsuperscript{2}] & [rad]  \\ \hline
		1 & $\pm 2.9234$ & $\pm 1.45 $ & $\pm 9 $ & $-0.7854 $ \\ \hline
		2 & $\pm 2.0508 $ & $\pm 1.45$ & $\pm 9 $ & $+2.0502 $ \\ \hline
		3 & $\pm 2.9234 $ & $\pm 1.45$ & $\pm 9 $ & $+2.0721 $ \\ \hline
		4 & $\pm 2.0508 $ & $\pm 1.45$ & $\pm 9 $ & $-1.6563 $ \\ \hline 
		5 & $\pm 2.9234 $ & $\pm 1.45$ & $\pm 9 $ & $-2.0893 $ \\ \hline
		6 & $\pm 2.0508 $ & $\pm 1.45$ & $\pm 9 $ & $+2.0342 $ \\ \hline
		7 & $\pm 3.0107 $ & $\pm 1.45$ & $\pm 9 $ & $ 0 $
	\end{tabularx}
\end{table}

\subsection{First set of experiments with the LBR iiwa robot}
\label{sec:4_results_exp_iiwa}
In the first set of experiments the LBR iiwa performs repeatedly the same tasks with the basic eSNS solver working at velocity, acceleration and torque level. In the first two cases the output of the solver is integrated to obtain a joint position reference, which is then provided to a low-level position controller; in the last case the output of the algorithm is directly fed to the joint actuators and an actual control loop is implemented. {\color{black}All the algorithms run as real-time module in a VxWorks\textsuperscript{\textregistered} (32 bit) environment on one of the four cores of an Intel Core\texttrademark i5-45705 (2.89 GHz) CPU with 432 MB of dedicated RAM. The control cycle time is $1$ ms.}
\begin{figure}[t!]
	\centering
	\includegraphics[width=0.49\textwidth]{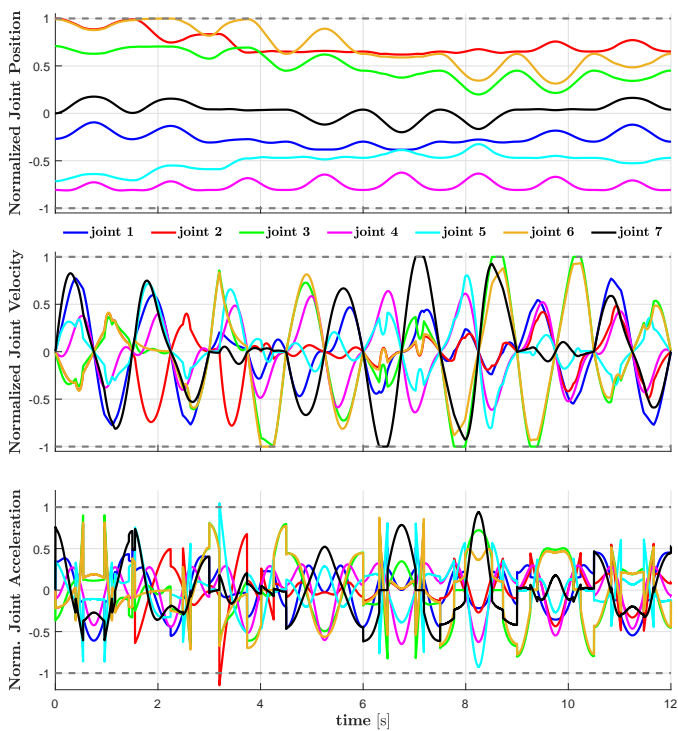}
	\caption{Normalized joint position, velocity and acceleration produced by the velocity-based eSNS solver in the experiments of Sect. \ref{sec:4_results_exp_iiwa}. }
	\label{fig:experiment1jointPosVelAcc_vel}
\end{figure}
\begin{figure}[t!]
	\centering
	\includegraphics[width=0.49\textwidth]{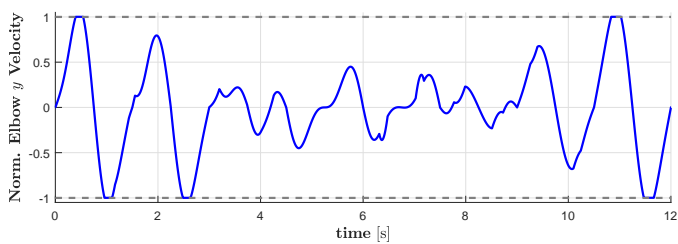}
	\caption{Normalized elbow velocity in $y$-direction produced by the velocity-based eSNS solver in the experiments of Sect. \ref{sec:4_results_exp_iiwa}.}
	\label{fig:experiment1Elbow_vel}
\end{figure}

The first task is to track a given desired Cartesian trajectory $(\cvec{x}_d,\cvecdot{x}_d,\cvecddot{x}_d)$ with the center point of the robot end-effector (task space equality constraint). Figure \ref{fig:iiwaMotion} shows the Cartesian path used for the experiments. The robot is commanded to move on each segment of the star, returning every time to the center point, following a sinusoidal velocity profile. The total planned time is 12 seconds (1.5 seconds per segment). The initial end-effector orientation must be kept along the entire trajectory. Therefore, both Cartesian position and orientation of the robot are controlled and only one redundant DOF is available.

The second task is to keep the velocity of the center point of the robot elbow along the direction of the $y$-axis below the limit of $0.35\,$m/s (task space inequality constraint). 
Additional limitations on joint position, velocity and acceleration are considered, which can be seen in Tab. \ref{tab:jointPosLim}. These limitations will originate a set of additional (joint space) inequality constraints for the solver. 

\begin{figure}[t!]
	\centering
	\includegraphics[width=0.49\textwidth]{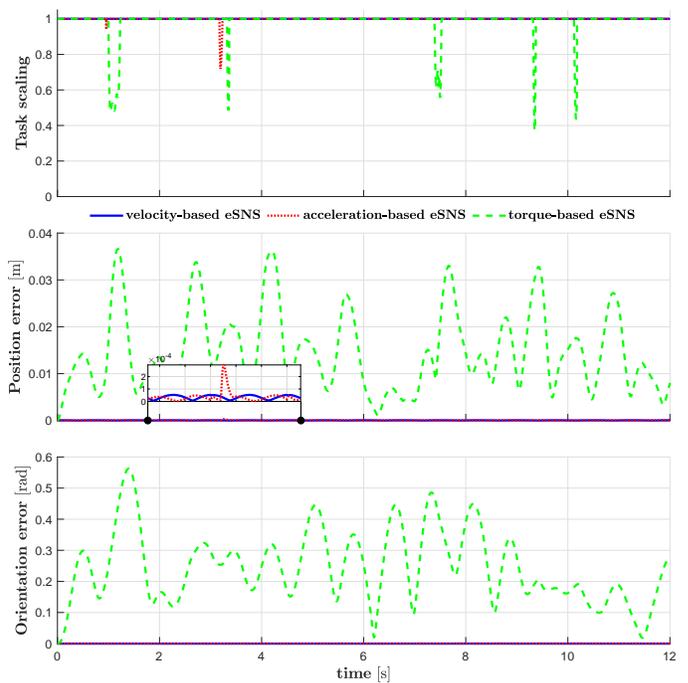}
	\caption{Task scaling factor (above) and Cartesian tracking error for the experiments of Sect. \ref{sec:4_results_exp_iiwa}; position error (in norm) is presented in the center; orientation error, reported as the angle extracted from the quaternion error, is reported below.}
	\label{fig:experiment1scFactCartError}
\end{figure}

The tasks are handled using one level of priority, meaning that the inequalities regarding joint and elbow limitations are stacked and treated as one set of constraints. The initial configuration of the robot, which can be seen in the top-left corner of Fig. \ref{fig:iiwaMotion}, places the second and the sixth joint very close to their respective upper position limit (see Tab. \ref{tab:jointPosLim}). 
\begin{figure}[t!]
	\centering
	\includegraphics[width=0.49\textwidth]{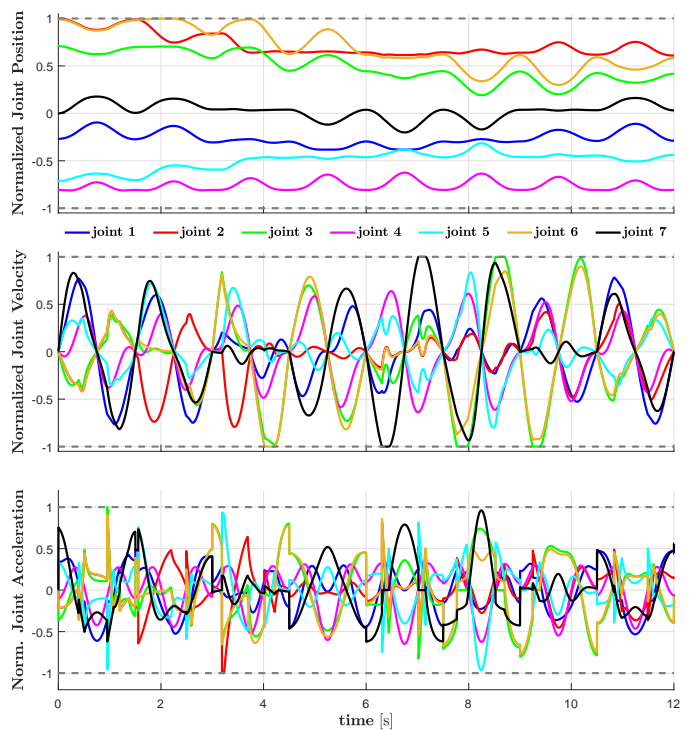}
	\caption{Normalized joint position, velocity and acceleration produced by the acceleration-based eSNS solver in the experiments of Sect. \ref{sec:4_results_exp_iiwa}.}
	\label{fig:experiment1jointPosVelAcc_acc}
\end{figure}
\begin{figure}[t!]
	\centering
	\includegraphics[width=0.49\textwidth]{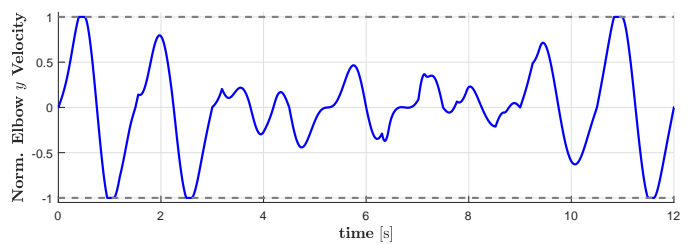}
	\caption{Normalized elbow velocity in $y$-direction produced by the acceleration-based eSNS solver in the experiments of Sect. \ref{sec:4_results_exp_iiwa}.}
	\label{fig:experiment1Elbow_acc}
\end{figure}
The velocity-based eSNS solver uses the velocity bounds in \bracketref{eq:velbounds} with $\cmat{K}=10\cmat{I}$, $\cmat{H}=\cmat{M}$ and $\cvec{u}_r = \cvec{0}$. The task reference velocity is defined as in \bracketref{eq:clik_vel}, with $\cmat{K}=50\cmat{I}$. The output joint velocities are shown in Fig. \ref{fig:experiment1jointPosVelAcc_vel} and have been integrated and differentiated, so as to obtain the corresponding joint position and acceleration. For the sake of clarity, all the values are reported after a normalization that brings the allowed ranges defined in Tab. \ref{tab:jointPosLim} to the interval $[-1,1]$. Figure \ref{fig:experiment1Elbow_vel} shows the resulting elbow velocity along the direction of the $y$-axis. Also this signal is reported normalized with respect to to its allowed range. The intensive occurrence of saturation can be easily identified in Fig. \ref{fig:experiment1jointPosVelAcc_vel} and Fig. \ref{fig:experiment1Elbow_vel}, proving the effectiveness of the proposed algorithm in respecting the hard bounds imposed by the inequality constraints in both joint and task space. Furthermore, the task scaling factor remains constant and equal to $1$, indicating that the end-effector task remains feasible during the entire motion (Fig. \ref{fig:experiment1scFactCartError}). On the other hand, Figure \ref{fig:experiment1jointPosVelAcc_vel} also shows violation of the joint acceleration limits. The fulfillment of such limits cannot be guaranteed by the velocity bounds in \bracketref{eq:velbounds}.

\begin{figure}[t!]
	\centering
	\includegraphics[width=0.49\textwidth]{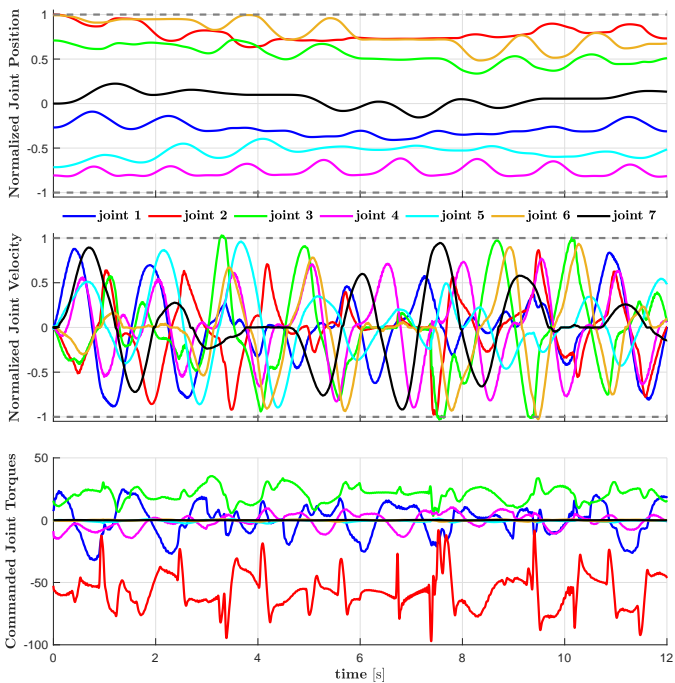}
	\caption{{\color{black}Joint motion produced by the torque-based eSNS solver in the experiments of Sect. \ref{sec:4_results_exp_iiwa}; since a measure of the joint acceleration is not available, only (normalized) joint position and velocity are reported alongside the commanded joint torques, which are the actual output of the solver.}}
	\label{fig:experiment1jointPosAndVel_trq}
\end{figure}
\begin{figure}[t]
	\centering
	\includegraphics[width=0.49\textwidth]{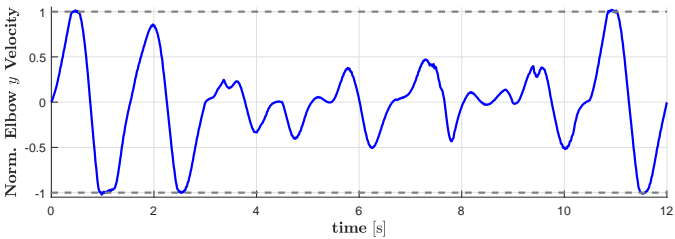}
	\caption{Normalized elbow velocity in $y$-direction produced by the torque-based eSNS solver in the experiments of Sect. \ref{sec:4_results_exp_iiwa}.}
	\label{fig:experiment1Elbow_trq}
\end{figure}

The acceleration-based eSNS solver uses $\cmat{H} = \cmat{M}$ and the acceleration bounds in \bracketref{eq:accbounds}, with $\cmat{D}=40\cmat{I}$ and $\cmat{K}_1 = 10$. The reference task acceleration is defined as in \bracketref{eq:clik_acc}, with $\cmat{K}=400\cmat{I}$ and $\cmat{D}=40\cmat{I}$. Furthermore, the choice $\cvec{u}_r = -k_{damp}\cvec{\dot{q}}(t)$ (with $k_{damp}$ a positive gain set to 20) is used to damp null space motions. The generated joint motion is shown in Fig. \ref{fig:experiment1jointPosVelAcc_acc}; compared to Fig. \ref{fig:experiment1jointPosVelAcc_vel}, the constraint on the maximum joint acceleration is now fulfilled. On the other hand, it is more difficult for the algorithm to find a feasible solution throughout the Cartesian end-effector trajectory and a task scaling smaller than one is observed in some cases (Fig. \ref{fig:experiment1scFactCartError}). The elbow velocity along the $y$-direction is also reported (Fig. \ref{fig:experiment1Elbow_acc}).

The torque-based eSNS solver uses the same task reference and bounds as the acceleration controller. Damping of null space motion is again included, this time with the choice $\cvec{u}_r =-\cmat{M}(k_{damp}\cvec{\dot{q}}(t))$, with $k_{damp}=20$. The choice $\cmat{H} = \cmat{M}^{-1}$ should produce the same joint motion as the acceleration-based solver. However, inaccuracy in the estimation of the complete robot dynamic model results in a different motion (see Fig. \ref{fig:experiment1jointPosAndVel_trq}); for the same reason, small inaccuracies can be noticed on the joint velocity saturation, as well as on the saturation of the elbow velocity along the $y$-direction (see Fig. \ref{fig:experiment1Elbow_trq}). A different trend of the task scaling factor is also observed in Fig. \ref{fig:experiment1scFactCartError}. {\color{black} For the sake of completeness, Fig. \ref{fig:experiment1jointPosAndVel_trq} also reports the trend of the commanded joint torques, which are the actual output of the solver.}

The tracking of the Cartesian end-effector trajectory is also poorer when the robot is torque-controlled (Fig. \ref{fig:experiment1scFactCartError}). In the other cases, the tracking error is considerably smaller and increases only when a task scaling $s<1$ is observed, as visible in the enlargement of the center plot.

\subsection{Second set of Experiments with the LBR iiwa robot}
\label{sec:4_results_sim_iiwa}
\begin{figure}[t!]
	\centering
	\includegraphics[width=0.49\textwidth]{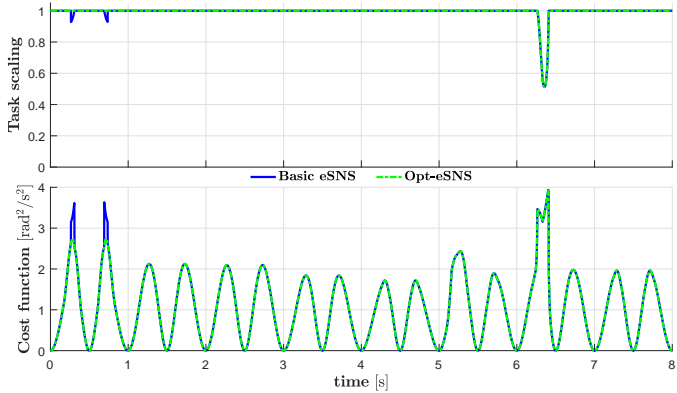}
	\caption{Task scaling factors (above) and cost function values (below) produced by the different eSNS variants in the experiments of Sect. \ref{sec:4_results_sim_iiwa}}
	\label{fig:experiment2scFactAndCost}
\end{figure}

\begin{figure}[t]
	\centering
	\includegraphics[width=0.49\textwidth]{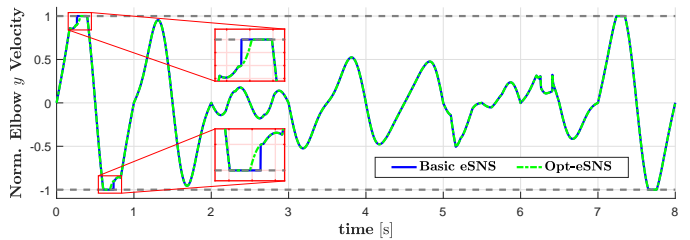}
	\caption{Normalized elbow velocity in $y$-direction produced by the different eSNS variants in the experiments of Sect. \ref{sec:4_results_sim_iiwa}}
	\label{fig:experiment2elbVel}
\end{figure}

The main goal of the second set of experiments is to highlight the benefits of the Opt-eSNS presented in Sect. \ref{sec:3_opteSNS}, compared to the basic eSNS algorithm. {\color{black} The experimental setup presents the same computing hardware and control cycle time as Sect. \ref{sec:4_results_exp_iiwa}, while the task definition has been slightly modified.} Indeed, the LBR iiwa is commanded again to track the Cartesian trajectory described in Sect. \ref{sec:4_results_exp_iiwa} with its end-effector center point. However, this time the orientation is not constrained, leaving four redundant DOFs available. Moreover, the total planned time is reduced to 8 seconds (1 second per each star segment) and the initial joint configuration is changed to $\cvec{q}_0 = [ -0.9 \; 2 \; 2.1721 \; -1.8055 \; -2.0893 \; 1.9834 \; 0]^T\,$rad. The same inequality constraints (elbow velocity along the direction of the $y$-axis and joint limits) of Sect. \ref{sec:4_results_exp_iiwa} are instead considered. All the tasks (equality and inequality constraints) are handled using one level of priority and solved using the velocity-based solver with $\cmat{H} = \cmat{I}$ and $\cvec{u}_r=\cvec{0}$. Velocity bounds are computed according to \bracketref{eq:velbounds} using $\cmat{K}=10\cmat{I}$, whereas task references are computed according to \bracketref{eq:clik_vel}, with $\cmat{K}=50\cmat{I}$. This specific setup has been chosen because it clearly shows the non optimality of the solution returned by the basic eSNS. Indeed, the basic eSNS and Opt-eSNS show quite different results when used to perform the assigned tasks. {\color{black} The disparity between the two solutions comes from the ability of the Opt-eSNS of removing constraints from the (optimal) saturation set.} Figure \ref{fig:experiment2scFactAndCost} reports the task scaling factors and cost function ($\frac{1}{2} \left( \cvec{u}-\cvec{u}_r \right)^T \cmat{H} \left( \cvec{u}-\cvec{u}_r\right)$) values over time obtained by the two algorithms. The difference in the results is especially visible in the first second of motion and it is highly reflected on the trend of the elbow velocity along the $y$ direction. As visible in Fig. \ref{fig:experiment2elbVel}, the non-optimal saturation sets obtained through the basic eSNS lead to sudden variations in the first second of motion, whereas the Opt-eSNS returns smoother motions.
\begin{figure}[htbp]
	\centering
	\includegraphics[width=0.49\textwidth]{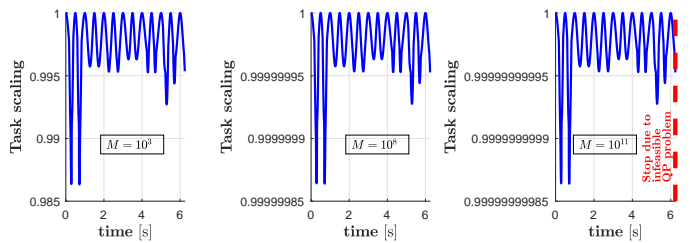}
	\caption{Task scaling factor produced by the state-of-the-art QP solver as $M$ changes in the experiments of Sect. \ref{sec:4_results_sim_iiwa}. For a meaningful comparison, only the first 6.2 seconds of motion are considered in every case.}
	\label{fig:experiment2scFactM}
\end{figure}
To validate the optimality of the results of the Opt-eSNS, the obtained solution has been compared with the one returned by a state-of-the-art QP solver (\textit{qpOASES} from \cite{ferreau2014qpoases}), operating on the same set of constraints and the cost function in \bracketref{eq:opt_prob_k_Mscale}. The average task scaling factor and the average value of $\frac{1}{2} \left( \cvec{u}-\cvec{u}_r \right)^T \cmat{H} \left( \cvec{u}-\cvec{u}_r\right)$ (which in this special case coincides with the joint velocity norm function) are reported in Tab. \ref{tab:averageVals} for the different algorithms used in the experiments of this section. It can be noticed that the Opt-eSNS provides both a higher average task scaling factor and a lower cost function value in the considered experiment, compared to the basic eSNS algorithm.

\begin{table}[t!]
	\centering
	\caption{Average task scaling factor and cost function value for the experiments of Sect. \ref{sec:4_results_sim_iiwa}.}
	\label{tab:averageVals}
	\begin{tabularx}{0.49\textwidth}{c|c|c}
		Algorithm & Task scaling factor & $\frac{1}{2} \left( \cvec{u}-\cvec{u}_r \right)^T \cmat{H} \left( \cvec{u}-\cvec{u}_r\right)$ \\ 
	          & $s$                 & [rad\textsuperscript{2}/s\textsuperscript{2}]                      \\ \hline
		Basic e-SNS & $0.9939$ & $1.0533$ \\ \hline
		Opt e-SNS   & $0.9944$ & $1.0459$ \\ \hline
		qpOases ($M=10^3$)   & $0.9917$ & $1.0443$ \\ \hline
		qpOases ($M=10^8$)   & $0.9944$ & $1.0459$ 
	\end{tabularx}
\end{table}
Finally, the importance of choosing a good value for the parameter $M$ when using a state-of-the-art QP solver should be remarked, as opposed to the absence of such parameter in the Opt-eSNS. Values of $M$ that are too small could lead to non optimal solutions. In our experiments, e.g., a value of $M=10^3$ returned worse results than the basic eSNS in terms of average task scaling factor (see Tab. \ref{tab:averageVals}). On the other hand, bigger values of $M$ may lead to numerical instability and, thus, to infeasible QP problems (Fig. \ref{fig:experiment2scFactM}).

\subsection{Simulations with the mobile dual-arm system}
\label{sec:4_results_sim_diiwa}
\begin{figure*}[t!]
	\centering
	\includegraphics[width=0.99\textwidth]{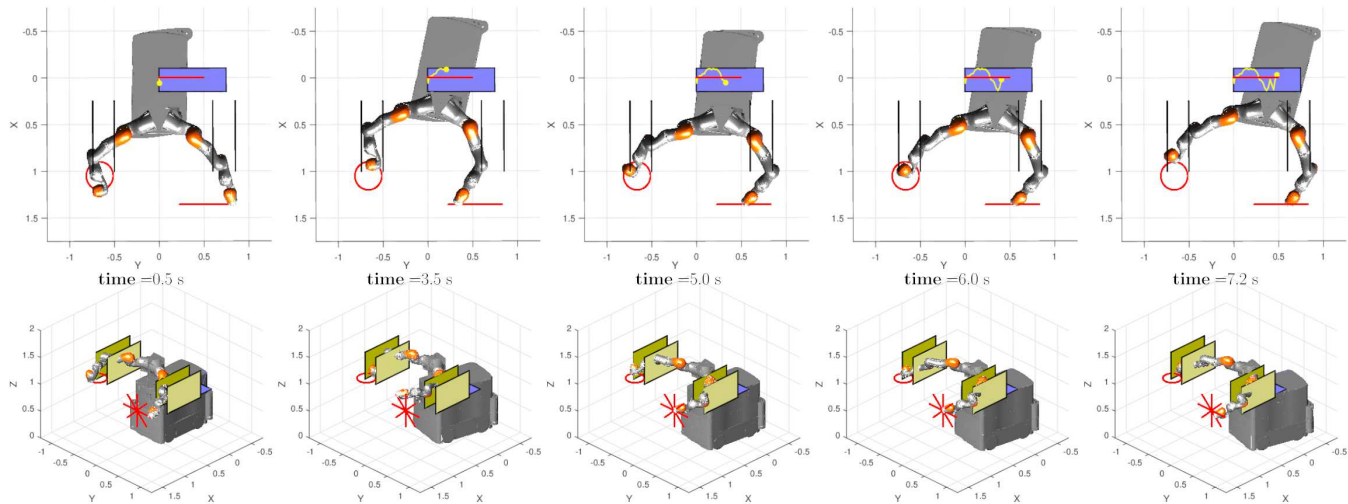}
	\caption{Motion sequence of the mobile dual-arm robot in the simulation of Sect. \ref{sec:4_results_sim_diiwa}. The top row shows a top view of the motion, whereas the bottom row offers a side view.}
	\label{fig:experiment3diiwa_seq}
\end{figure*}

This section presents the results of a set of simulations involving a mobile dual arm manipulator with 17 DOFs (Fig.  \ref{fig:experiment3diiwa_seq}): the mobile base is equipped with omni-directional wheels and it is therefore modeled as a sequence of two prismatic joints and a revolute one, all located at the center of the base; additionally, two LBR iiwa robots are mounted on top of the mobile base. The main objective of the simulations is to prove the effectiveness of the proposed algorithms in handling multiple levels of priority. Moreover, the performance of the different eSNS variants is evaluated. The simulations are carried out in Matlab\textsuperscript{\textregistered} environment running on an Intel\textsuperscript{\textregistered} Core\texttrademark i7-8850H (2.60GHz) CPU with 16 GB of RAM.
\begin{table}[t!]
	\centering
	\caption{Limits and initial positions for the joints of the mobile base and the elbow center points for the simulations of Sect. \ref{sec:4_results_sim_diiwa}.}
	\label{tab:diiwaPosLim}
	\begin{tabularx}{0.49\textwidth}{m{0.09\textwidth}|m{0.065\textwidth}|m{0.065\textwidth}|m{0.07\textwidth}|m{0.075\textwidth}}
		 & Lower Position Limit $\cvecbardown{p}$  & Upper Position Limit $\cvecbarup{p}$ & Velocity \newline Limit $\cvecbardown{v},\cvecbarup{v}$  & Initial \newline Position\\ \hline
		Mobile Base Joint 1    & $-0.1\,$m   & $0.15\,$m  & $\pm 0.5\,$m/s   & $0\,$m     \\ \hline
		Mobile Base Joint 2    & $0\,$m      & $0.75\,$m  & $\pm 0.5\,$m/s   & $0\,$m     \\ \hline
		Mobile Base Joint 3    & $-0.5\,$rad & $0.5\,$rad & $\pm 0.5\,$rad/s & $0\,$rad   \\ \hline
		Left Elbow $y$ Coordinate  & $0.6\,$m    & $0.85\,$m  & $+\inf\,$m/s     & $0.6216\,$m\\ \hline
		Right Elbow $y$ Coordinate & $-0.75\,$m  & $-0.5\,$m  & $-\inf\,$m/s     & $-0.7127\,$m
	\end{tabularx}
\end{table}
\begin{table}[t!]
	\centering
	\caption{Task hierarchy for the simulations of Sect. \ref{sec:4_results_sim_diiwa}.}
	\label{tab:diiwaTaskHierarchy}
	\begin{tabularx}{0.49\textwidth}{m{0.04\textwidth}|m{0.18\textwidth}|m{0.18\textwidth}}
		Priority Level & Inequality Constraints  & Equality Constraints \\ \hline
		\quad 1    & Joint position and velocity limits (dim. $17$) & Desired trajectory for the mobile base $y$ coordinate (dim. 1)       \\ \hline
		\quad 2    & Position limits on left elbow $y$ coordinate (dim. $1$) & Desired trajectory for the left end effector center point (dim. 3)       \\ \hline
		\quad 3    & Position limits on left elbow $y$ coordinate (dim. $1$) & Desired trajectory for the right end effector center point (dim. 3)       
    \end{tabularx}
\end{table}

{\color{black} The tasks to execute are distributed on three levels of priority. At the first level, the center point of the mobile base (yellow point in Fig. \ref{fig:experiment3diiwa_seq}) is required to track a desired 1D trajectory along the direction of the $Y$ axis. The trajectory starts at $y=0$ and ends in $y=0.5$, following a trapezoidal velocity profile. The total planned time is 8 seconds. The top row of Fig. \ref{fig:experiment3diiwa_seq} shows the desired $y$ trajectory (drawn for a fixed value $x=0$) as a red segment. Additionally, limitations on the position and the velocity of each joint are considered: the limit values used for the mobile base are reported in Tab. \ref{tab:diiwaPosLim}, whereas the limits from Tab. \ref{tab:jointPosLim} are used for the two robotic arms. The position limits of the first two joints are represented by the blue rectangle in Fig. \ref{fig:experiment3diiwa_seq}.
At the second priority level, the left arm is required to track a desired 3D positional Cartesian trajectory with its end-effector center point. The trajectory consists in the star-like path with sinusoidal velocity profile introduced in Sect. \ref{sec:4_results_exp_iiwa}. The total planned time is again 8 seconds (1 second per star-segment). Moreover, limitations on the position along the $y$-axis are imposed on the left elbow center point. These limitations are represented by the yellow planes in Fig. \ref{fig:experiment3diiwa_seq}, while the numeric values of the lower and upper position limit are reported in Tab. \ref{tab:diiwaPosLim}.

On the third level of priority, a 3D positional Cartesian trajectory is assigned to the end-effector center point of the right arm. The trajectory consists of a circular path defined in the $XY$-plane (center $c_0 = [1.0496 \ -0.6635 \ 1.0690]\,$m, radius $r_0 = 0.15\,$m) with a trapezoidal velocity profile. The planned time to cover the circular path is also 8 seconds.  Moreover, limitations on the position along the $y$-axis are imposed on the right elbow center point. As for the left arm, such limitations are represented by yellow planes in Fig. \ref{fig:experiment3diiwa_seq}, while the numeric values of the lower and upper position limits are reported in Tab. \ref{tab:diiwaPosLim}. An overview of the considered set of tasks is given in Table \ref{tab:diiwaTaskHierarchy}.

The tasks are solved by using the basic eSNS, operating at velocity level with $\cmat{H}=\cmat{I}$ and $\cvec{u}_r=\cvec{0}$. All the velocity bounds are computed according to \bracketref{eq:velbounds}, with $\cmat{K} = 10\cmat{I}$, whereas all task references are computed according to \bracketref{eq:clik_vel} using $\cmat{K}=100\cmat{I}$. The initial joint configuration of the mobile platform is reported in Tab. \ref{tab:diiwaPosLim}, while left and right arms start from $\cvec{q}_{0,l}=[-1.2 \; -0.57 \; 0 \; 0.78 \; 0 \; 0 \; 0]^T\,$rad and $\cvec{q}_{0,r}=[-1.9 \; 0.27 \; 0 \; -0.98 \; 1.8 \; -1.2 \; 0]^T\,$rad, respectively. The control cycle time used in the simulation is 1 ms.

The trend of the joint position and velocity over time is reported in Fig. \ref{fig:experiment3diiwa_posvel}, whereas Fig. \ref{fig:experiment3diiwa_elbow} shows the normalized position along the $y$-axis of the elbow center points; all the quantities are reported normalized with respect to their respective admissible range of motion. Finally, Fig. \ref{fig:experiment3scFactCartError} reports the trend of the scale factors and the tracking errors for each priority level.

\begin{figure*}[t!]
	\centering
	\includegraphics[width=0.99\textwidth]{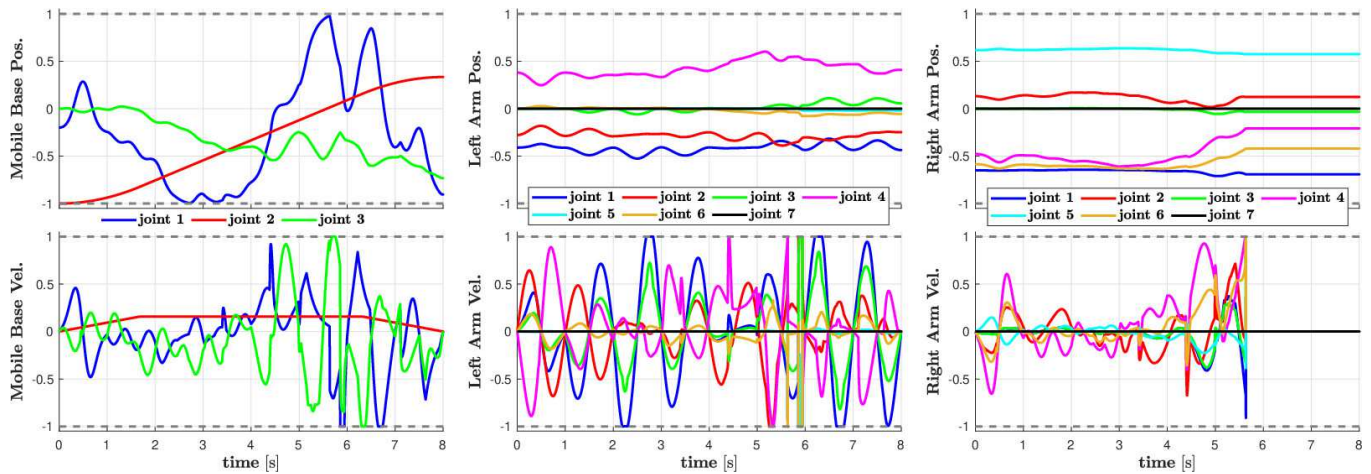}
	\caption{Normalized joint position and velocity produced by the velocity-based eSNS solver in the  Simulation of Sect. \ref{sec:4_results_sim_diiwa}. Joint motion is reported separately for the mobile base (left), the left arm (center) and the right arm (right).}
	\label{fig:experiment3diiwa_posvel}
\end{figure*}
\begin{figure}[t!]
	\centering
	\includegraphics[width=0.49\textwidth]{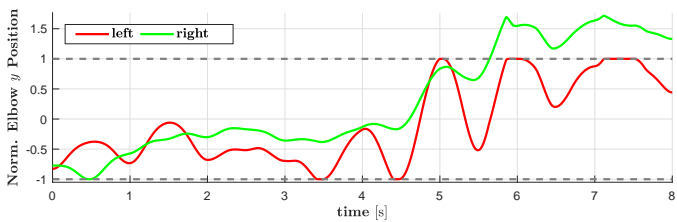}
	\caption{Normalized elbow positions in $y$-direction produced by the velocity-based eSNS solver in the Simulation of Sect. \ref{sec:4_results_sim_diiwa}.}
	\label{fig:experiment3diiwa_elbow}
\end{figure}
\begin{figure}[t!]
	\centering
	\includegraphics[width=0.49\textwidth]{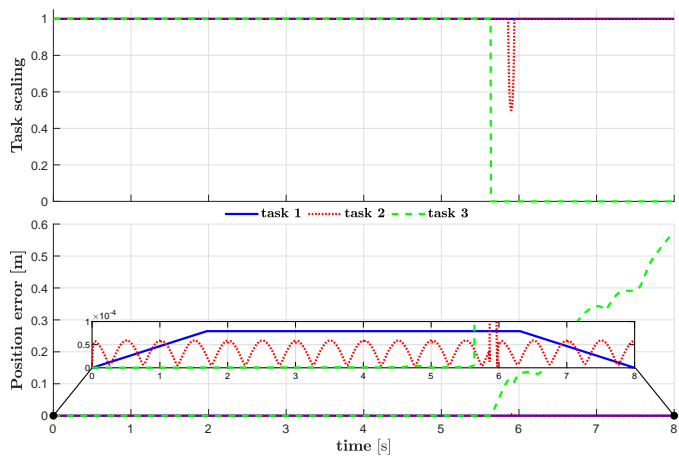}
	\caption{Task scaling factors (above) and norm of Cartesian tracking position errors (below) produced by the velocity-based eSNS solver in the simulation of Sect. \ref{sec:4_results_sim_diiwa}.}
	\label{fig:experiment3scFactCartError}
\end{figure}

The intensive saturation of different joint and task space variables can be noticed in Fig. \ref{fig:experiment3diiwa_posvel} and \ref{fig:experiment3diiwa_elbow}. Having the highest priority, joint limits are never violated. Satisfying the equality constraint of the first priority level is also possible, as shown by the constant task scaling factor (equal to $1$) and the small tracking error in Fig. \ref{fig:experiment3scFactCartError}. On the other hand, the demanding task specification produces task scaling on the second (from time $t=5.86$ s until $t=5.940$ s) and third (from $t=5.634$) priority level. More specifically, the task scaling factor of third task rapidly decreases to zero as the related constraints become incompatible with the task hierarchy. As a consequence, the task is completely sacrificed, leading to violation of the right elbow position limit (Fig. \ref{fig:experiment3diiwa_elbow}) and an increase of the error in tracking the circular trajectory (Fig. \ref{fig:experiment3scFactCartError}). It is worth noticing that at each priority level the tracking error remains limited throughout the entire motion and increases only when the corresponding task scaling factor becomes smaller than $1$. This can be easily noticed in the enlargement of the bottom plot in Fig. \ref{fig:experiment3scFactCartError}.}
\begin{figure*}[t!]
	\centering
	\includegraphics[width=0.99\textwidth]{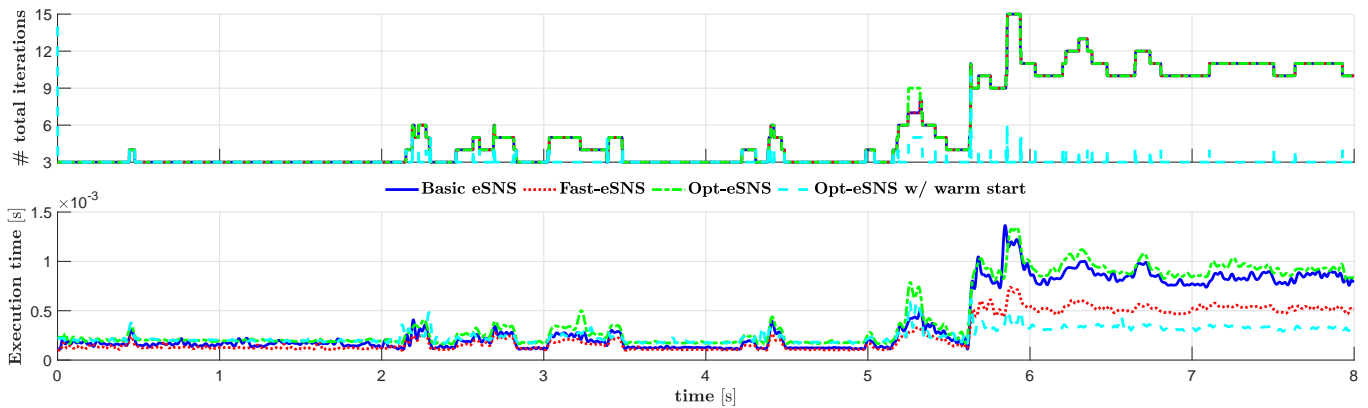}
	\caption{Number of total iterations (top plot) and execution time (bottom plot) of the different eSNS variants for the simulations of Sect. \ref{sec:4_results_sim_diiwa}.}
	\label{fig:experiment3NItTimes}
\end{figure*}
\begin{figure}[t!]
	\centering
	\includegraphics[width=0.49\textwidth]{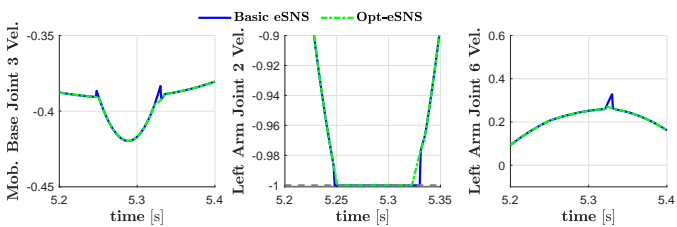}
	\caption{Comparison between the joint velocity obtained with the Opt-eSNS and the basic eSNS in the simulations of Sect. \ref{sec:4_results_sim_diiwa}; for brevity, only the joints presenting significant differences are reported.}
	\label{fig:experiment3OptDiff}
\end{figure}
\begin{figure}[t!]
	\centering
	\includegraphics[width=0.49\textwidth]{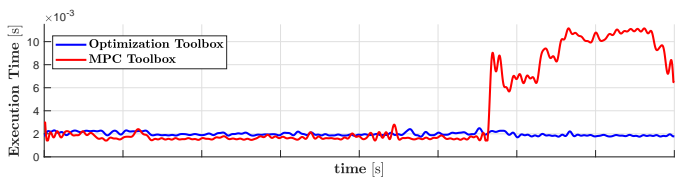}
	\caption{Execution time of the state-of-the-art QP solvers in the simulations of Sect. \ref{sec:4_results_sim_diiwa}.}
	\label{fig:experiment3QPSolvTime}
\end{figure}

{\color{black}To evaluate the performance of the different eSNS variants, the same simulation has been repeated using the Fast-eSNS, the Opt-eSNS and the Opt-eSNS with warm start. The joint motion produced by the Fast-eSNS is, as expected, identical to the one obtained with the basic eSNS and it is therefore not reported. The main advantage of this variant lies in its computational efficiency. This can be appreciated in Fig. \ref{fig:experiment3NItTimes}, where the total number of iterations and the execution time of each eSNS variant are reported. It can be easily noticed that the Fast-eSNS always provides faster computation compared to the basic eSNS, with the difference on the execution time that is as significant as the number of total iterations increases. The Opt-eSNS returns, in light of the optimality check performed inside the algorithm, slightly different solutions. This can be directly noticed in Fig. \ref{fig:experiment3OptDiff}, which reports the velocity of the joints that present a more significant difference, compared to the solution obtained with the basic eSNS. Moreover, differences can also be noticed in the number of total iterations performed by the algorithm (Fig. \ref{fig:experiment3NItTimes}), which is in some cases higher than the basic eSNS due to the removal of constraints from the saturation set. Figure \ref{fig:experiment3NItTimes} also shows that the Opt-eSNS presents computation times comparable to the basic eSNS, proving that the computational burden added by the optimality check is very limited. Since (as expected) the saturation sets do not change so much and/or so often between two consecutive instants of time, the total number of iterations is dramatically reduced when using the Opt-eSNS with warm start. As a consequence, the execution time is sometimes comparable or even lower than the Fast-eSNS. This is reasonable, when the difference in terms of the number of total iterations required by the two algorithms becomes significant. It also provides a good insight on how much the performance of the eSNS algorithm could be sped up, if the efficient computation of the Fast-eSNS would be combined with the warm start technique from the Opt-eSNS. Such further development is left as future work.

Finally, the obtained execution times have been compared with the ones returned by state-of-the-art QP solvers working on optimization problems in the form \bracketref{eq:opt_prob_k_Mscale}. For a meaningful comparison, we have selected solvers that are implemented as pure MATLAB\textsuperscript{\textregistered} code as the algorithms used in this section. In particular, the Optimization Toolbox\texttrademark and the Model Predictive Control Toolbox\texttrademark offer two different active-set methods with warm start. The obtained execution times are reported in Fig. \ref{fig:experiment3QPSolvTime}. A direct comparison with Fig. \ref{fig:experiment3NItTimes} shows that the state-of-the-art QP solvers present significantly higher computation times than all presented eSNS variants throughout the motion.}

\section{Conclusion}
This paper has presented a general framework for the motion control of redundant robots. Thanks to a unified formulation of the redundancy resolution problem, the framework can arbitrary resolve redundancy at velocity, acceleration or torque level. Thus, both kinematic and torque control are possible, providing the framework with high versatility with respect to the different application domains and robot interfaces. 

The unified formulation has led to the definition of a generalized control problem, which retains a certain number of essential features. First of all, both equality and inequality constraints can be defined in any task space and with arbitrary priority, allowing high flexibility in the task specification. Methods to define task velocity/acceleration references (equality constraints) and bounds (inequality constraints) have also been discussed. In particular, a novel shaping of the task velocity and acceleration bounds has been introduced, which simplifies and at the same time extends existing solutions in the literature. Moreover, the unified control problem considers the minimization of an arbitrary weighted control effort and an additional input vector used to control possibly remaining redundant DOFs.

A unified formulation also allows for the design of a single redundancy resolution algorithm. This paper has presented a novel approach, i.e., the eSNS method, obtained by extending the existing SNS algorithm from the literature. The eSNS shares the same structure of SNS and retains all its features: strict prioritization of the constraints according to the specified hierarchy of tasks, handling of inequality constraints via saturation sets, integrated (multitask) scaling technique to extend the execution of infeasible tasks in time while preserving task directions. However, the application to the generalized control problem and the handling of all its features discussed above have required significant extensions to the method, as thoroughly detailed in Sect. \ref{sec:3_extendedSNS}. Additional algorithms have also been presented, namely the Fast-eSNS and Opt-eSNS, which tackle important aspects such as computation efficiency and optimality of the solution.

The effectiveness of the eSNS in solving the generalized control problem has been proved through simulations and experiments with different kinematic structures in Sect. \ref{sec:4_results}. The optimality of the solution obtained via the Opt-eSNS has been validated by comparing the results with the ones obtained by a state-of-the-art QP solver. The results of Sect. \ref{sec:4_results} have also shown a significant decrease in the computation time when the Opt-eSNS is warm-started, as well as when the Fast-eSNS is employed. A combination of these two variants should bring even faster computation times while retaining optimal control inputs. Such a solution will be investigated in the future. {\color{black} Another possible computation speed-up might also be obtained by passing the obtained saturation set from one level of priority to the next one similarly to \cite{escande2014hierarchical}. Such a solution could be investigated and compared to the Opt-eSNS with warm start.}

{\color{black}Finally, the performance of the torque-based eSNS during physical Human-Robot Interaction (pHRI) should be evaluated. In such scenario, special attention should be put in the consideration of external forces and on the choice of the matrix $\cmat{H}$ to avoid injection of active energy in the closed-loop system during interaction in the presence of saturation. An approach going in this direction can be found in \cite{osorio2022}.}

\printbibliography

@inproceedings{kuhn1951tucker,
  title={Nonlinear programming},
  author={Kuhn, HW and Tucker, AW},
  booktitle={Proc. 2nd Berkeley Symposium on Mathematical Statistics and Probability},
  pages={481--492},
  year={1951}
}

@article{greville1960some,
  title={Some applications of the pseudoinverse of a matrix},
  author={Greville, TNE},
  journal={SIAM review},
  volume={2},
  number={1},
  pages={15--22},
  year={1960},
  publisher={SIAM}
}

@article{Whitney.1969,
  title={Resolved motion rate control of manipulators and human prostheses},
  author={Whitney, Daniel E},
  journal={IEEE Transactions on man-machine systems},
  volume={10},
  number={2},
  pages={47--53},
  year={1969},
  publisher={IEEE}
}

@article{liegeois1977automatic,
  title={Automatic supervisory control of the configuration and behavior of multibody mechanisms},
  author={Liegeois, Alain and others},
  journal={IEEE transactions on systems, man, and cybernetics},
  volume={7},
  number={12},
  pages={868--871},
  year={1977}
}

@inproceedings{hollerbach1983dynamic,
  title={Dynamic scaling of manipulator trajectories},
  author={Hollerbach, John M},
  booktitle={1983 American Control Conference},
  pages={752--756},
  year={1983},
  organization={IEEE}
}

@inproceedings{khatib1983dynamic,
  title={Dynamic control of manipulator in operational space},
  author={Khatib, Oussama},
  booktitle={Proc. 6th IFToMM World Congress on Theory of Machines and Mechanisms},
  pages={1128--1131},
  year={1983}
}

@article{balestrino1984robust,
  title={Robust control of robotic manipulators},
  author={Balestrino, Aldo and De Maria, Giuseppe and Sciavicco, Lorenzo},
  journal={IFAC Proceedings Volumes},
  volume={17},
  number={2},
  pages={2435--2440},
  year={1984},
  publisher={Elsevier}
}

@article{maciejewski1985obstacle,
  title={Obstacle avoidance for kinematically redundant manipulators in dynamically varying environments},
  author={Maciejewski, Anthony A and Klein, Charles A},
  journal={The international journal of robotics research},
  volume={4},
  number={3},
  pages={109--117},
  year={1985},
  publisher={Sage Publications Sage CA: Thousand Oaks, CA}
}

@incollection{Khatib.1986,
  title={Real-time obstacle avoidance for manipulators and mobile robots},
  author={Khatib, Oussama},
  booktitle={Autonomous robot vehicles},
  pages={396--404},
  year={1986},
  publisher={Springer}
}

@article{guo1986recursive,
  title={A Recursive Algorithm for Computing the Weighted Moore-Penrose Inverse},
  author={Guo-rong, Wang and Yong-lin, Chen},
  journal={Journal of Computational Mathematics},
  pages={74--85},
  year={1986},
  publisher={JSTOR}
}

@inproceedings{Siciliano.1991,
  title={A general framework for managing multiple tasks in highly redundant robotic systems},
  author={Siciliano, Bruno and Slotine, J-JE},
  booktitle={Fifth International Conference on Advanced Robotics},
  pages={1211--1216},
  year={1991},
  organization={IEEE}
}

@article{chiacchio1991closed,
  title={Closed-loop inverse kinematics schemes for constrained redundant manipulators with task space augmentation and task priority strategy},
  author={Chiacchio, Pasquale and Chiaverini, Stefano and Sciavicco, Lorenzo and Siciliano, Bruno},
  journal={The International Journal of Robotics Research},
  volume={10},
  number={4},
  pages={410--425},
  year={1991},
  publisher={Sage Publications Sage CA: Thousand Oaks, CA}
}

@article{Siciliano.1992,
  title={Robot redundancy resolution at the acceleration level},
  author={De Luca, A and Oriolo, G and Siciliano, B},
  journal={Laboratory Robotics and Automation},
  volume={4},
  pages={97--97},
  year={1992},
  publisher={WILEY}
}

@article{chiacchio1995coping,
  title={Coping with joint velocity limits in first-order inverse kinematics algorithms: Analysis and real-time implementation},
  author={Chiacchio, Pasquale and Chiaverini, Stefano},
  journal={Robotica},
  volume={13},
  number={5},
  pages={515--519},
  year={1995},
  publisher={Cambridge University Press}
}

@ARTICLE{chiaverini97srtp,
  author={Chiaverini, S.},
  journal={IEEE Transactions on Robotics and Automation}, 
  title={Singularity-robust task-priority redundancy resolution for real-time kinematic control of robot manipulators}, 
  year={1997},
  volume={13},
  number={3},
  pages={398-410},
  doi={10.1109/70.585902}}

@article{antonelli2003new,
  title={A new on-line algorithm for inverse kinematics of robot manipulators ensuring path tracking capability under joint limits},
  author={Antonelli, Gianluca and Chiaverini, Stefano and Fusco, Giuseppe},
  journal={IEEE Transactions on Robotics and Automation},
  volume={19},
  number={1},
  pages={162--167},
  year={2003},
  publisher={IEEE}
}

@inproceedings{sentis2004prioritized,
  title={Prioritized multi-objective dynamics and control of robots in human environments},
  author={Sentis, Luis and Khatib, Oussama},
  booktitle={4th IEEE/RAS International Conference on Humanoid Robots, 2004.},
  volume={2},
  pages={764--780},
  year={2004},
  organization={IEEE}
}

@article{sentis2005synthesis,
  title={Synthesis of whole-body behaviors through hierarchical control of behavioral primitives},
  author={Sentis, Luis and Khatib, Oussama},
  journal={International Journal of Humanoid Robotics},
  volume={2},
  number={04},
  pages={505--518},
  year={2005},
  publisher={World Scientific}
}

@article{peters2008unifying,
  title={A unifying framework for robot control with redundant DOFs},
  author={Peters, Jan and Mistry, Michael and Udwadia, Firdaus and Nakanishi, Jun and Schaal, Stefan},
  journal={Autonomous Robots},
  volume={24},
  number={1},
  pages={1--12},
  year={2008},
  publisher={Springer}
}

@article{mansard2009unified,
  title={A unified approach to integrate unilateral constraints in the stack of tasks},
  author={Mansard, Nicolas and Khatib, Oussama and Kheddar, Abderrahmane},
  journal={IEEE Transactions on Robotics},
  volume={25},
  number={3},
  pages={670--685},
  year={2009},
  publisher={IEEE}
}

@article{kanoun2011kinematic,
  title={Kinematic control of redundant manipulators: Generalizing the task-priority framework to inequality task},
  author={Kanoun, Oussama and Lamiraux, Florent and Wieber, Pierre-Brice},
  journal={IEEE Transactions on Robotics},
  volume={27},
  number={4},
  pages={785--792},
  year={2011},
  publisher={IEEE}
}

@article{lee2012intermediate,
  title={Intermediate desired value approach for task transition of robots in kinematic control},
  author={Lee, Jaemin and Mansard, Nicolas and Park, Jaeheung},
  journal={IEEE Transactions on Robotics},
  volume={28},
  number={6},
  pages={1260--1277},
  year={2012},
  publisher={IEEE}
}

@article{rubrecht2012motion,
  title={Motion safety and constraints compatibility for multibody robots},
  author={Rubrecht, S{\'e}bastien and Padois, Vincent and Bidaud, Philippe and De Broissia, Michel and Simoes, Max Da Silva},
  journal={Autonomous Robots},
  volume={32},
  number={3},
  pages={333--349},
  year={2012},
  publisher={Springer}
}

@inproceedings{flacco2012prioritized,
  title={Prioritized multi-task motion control of redundant robots under hard joint constraints},
  author={Flacco, Fabrizio and De Luca, Alessandro and Khatib, Oussama},
  booktitle={2012 IEEE/RSJ International Conference on Intelligent Robots and Systems},
  pages={3970--3977},
  year={2012},
  organization={IEEE}
}

@inproceedings{flacco2013optimal,
  title={Optimal redundancy resolution with task scaling under hard bounds in the robot joint space},
  author={Flacco, Fabrizio and De Luca, Alessandro},
  booktitle={2013 IEEE International Conference on Robotics and Automation},
  pages={3969--3975},
  year={2013},
  organization={IEEE}
}

@inproceedings{flacco2013fast,
  title={Fast redundancy resolution for high-dimensional robots executing prioritized tasks under hard bounds in the joint space},
  author={Fiacco, Fabrizio and De Luca, Alessandro},
  booktitle={2013 IEEE/RSJ International Conference on Intelligent Robots and Systems},
  pages={2500--2506},
  year={2013},
  organization={IEEE}
}

@book{antonelli2014underwater,
  title={Underwater robots},
  author={Antonelli, Gianluca},
  volume={3},
  year={2014},
  publisher={Springer}
}

@article{escande2014hierarchical,
  title={Hierarchical quadratic programming: Fast online humanoid-robot motion generation},
  author={Escande, Adrien and Mansard, Nicolas and Wieber, Pierre-Brice},
  journal={The International Journal of Robotics Research},
  volume={33},
  number={7},
  pages={1006--1028},
  year={2014},
  publisher={SAGE Publications Sage UK: London, England}
}

@article{ferreau2014qpoases,
  title={qpOASES: A parametric active-set algorithm for quadratic programming},
  author={Ferreau, Hans Joachim and Kirches, Christian and Potschka, Andreas and Bock, Hans Georg and Diehl, Moritz},
  journal={Mathematical Programming Computation},
  volume={6},
  number={4},
  pages={327--363},
  year={2014},
  publisher={Springer}
}

@inproceedings{aertbelien2014etasl,
  title={eTaSL/eTC: A constraint-based task specification language and robot controller using expression graphs},
  author={Aertbeli{\"e}n, Erwin and De Schutter, Joris},
  booktitle={2014 IEEE/RSJ International Conference on Intelligent Robots and Systems},
  pages={1540--1546},
  year={2014},
  organization={IEEE}
}

@inproceedings{baizid2015experiments,
  title={Experiments on behavioral coordinated control of an unmanned aerial vehicle manipulator system},
  author={Baizid, Khelifa and Giglio, Gerardo and Pierri, Francesco and Trujillo, Miguel Angel and Antonelli, Gianluca and Caccavale, Fabrizio and Viguria, Antidio and Chiaverini, Stefano and Ollero, An{\'\i}bal},
  booktitle={2015 IEEE international conference on robotics and automation (ICRA)},
  pages={4680--4685},
  year={2015},
  organization={IEEE}
}

@article{flacco2015control,
  title={Control of redundant robots under hard joint constraints: Saturation in the null space},
  author={Flacco, Fabrizio and De Luca, Alessandro and Khatib, Oussama},
  journal={IEEE Transactions on Robotics},
  volume={31},
  number={3},
  pages={637--654},
  year={2015},
  publisher={IEEE}
}

@article{ott2015prioritized,
  title={Prioritized multi-task compliance control of redundant manipulators},
  author={Ott, Christian and Dietrich, Alexander and Albu-Sch{\"a}ffer, Alin},
  journal={Automatica},
  volume={53},
  pages={416--423},
  year={2015},
  publisher={Elsevier}
}

@article{liu2016generalized,
  title={Generalized hierarchical control},
  author={Liu, Mingxing and Tan, Yang and Padois, Vincent},
  journal={Autonomous Robots},
  volume={40},
  number={1},
  pages={17--31},
  year={2016},
  publisher={Springer}
}

@article{moe2016set,
  title={Set-based tasks within the singularity-robust multiple task-priority inverse kinematics framework: General formulation, stability analysis, and experimental results},
  author={Moe, Signe and Antonelli, Gianluca and Teel, Andrew R and Pettersen, Kristin Y and Schrimpf, Johannes},
  journal={Frontiers in Robotics and AI},
  volume={3},
  pages={16},
  year={2016},
  publisher={Frontiers}
}

@inproceedings{scheurer2016industrial,
  title={Industrial implementation of a multi-task redundancy resolution at velocity level for highly redundant mobile manipulators},
  author={Scheurer, Christian and Fiore, Mario Daniele and Sharma, Shashank and Natale, Ciro},
  booktitle={Proceedings of ISR 2016: 47st International Symposium on Robotics},
  pages={1--9},
  year={2016},
  organization={VDE}
}

@article{di2017comparison,
  title={A comparison of damped least squares algorithms for inverse kinematics of robot manipulators},
  author={Di Vito, Daniele and Natale, Ciro and Antonelli, Gianluca},
  journal={IFAC-PapersOnLine},
  volume={50},
  number={1},
  pages={6869--6874},
  year={2017},
  publisher={Elsevier}
}

@inproceedings{munoz2018operational,
  title={Operational Space Formulation Under Joint Constraints},
  author={Mu{\~n}oz Osorio, Juan D and Fiore, Mario D and Allmendinger, Felix},
  booktitle={International Design Engineering Technical Conferences and Computers and Information in Engineering Conference},
  volume={51814},
  year={2018},
  organization={American Society of Mechanical Engineers}
}

@inproceedings{hoffman2018multi,
  title={Multi-priority cartesian impedance control based on quadratic programming optimization},
  author={Hoffman, Enrico Mingo and Laurenzi, Arturo and Muratore, Luca and Tsagarakis, Nikos G and Caldwell, Darwin G},
  booktitle={2018 IEEE International Conference on Robotics and Automation (ICRA)},
  pages={309--315},
  year={2018},
  organization={IEEE}
}

@article{quiroz2019whole,
  title={Whole-body control with (self) collision avoidance using vector field inequalities},
  author={Quiroz-Oma{\~n}a, Juan Jos{\'e} and Adorno, Bruno Vilhena},
  journal={IEEE Robotics and Automation Letters},
  volume={4},
  number={4},
  pages={4048--4053},
  year={2019},
  publisher={IEEE}
}

@inproceedings{osorio2019physical,
  title={Physical human-robot interaction under joint and cartesian constraints},
  author={Osorio, Juan D Mu{\~n}oz and Allmendinger, Felix and Fiore, Mario D and Zimmermann, Uwe E and Ortmaier, Tobias},
  booktitle={2019 19th International Conference on Advanced Robotics (ICAR)},
  pages={185--191},
  year={2019},
  organization={IEEE}
}

@article{dietrich2019hierarchical,
  title={Hierarchical impedance-based tracking control of kinematically redundant robots},
  author={Dietrich, Alexander and Ott, Christian},
  journal={IEEE Transactions on Robotics},
  volume={36},
  number={1},
  pages={204--221},
  year={2019},
  publisher={IEEE}
}

@inproceedings{schettino2020geometrical,
  title={Geometrical Interpretation and Detection of Multiple Task Conflicts using a Coordinate Invariant Index},
  author={Schettino, Vincenzo and Fiore, Mario D and Pecorella, Claudia and Ficuciello, Fanny and Allmendinger, Felix and Lachner, Johannes and Stramigioli, Stefano and Siciliano, Bruno},
  booktitle={2020 IEEE/RSJ International Conference on Intelligent Robots and Systems (IROS)},
  pages={6613--6618},
  year={2020},
  organization={IEEE}
}

@inproceedings{ziese2020redundancy,
  title={Redundancy resolution under hard joint constraints: a generalized approach to rank updates},
  author={Ziese, Anton and Fiore, Mario D and Peters, Jan and Zimmermann, Uwe E and Adamy, J{\"u}rgen},
  booktitle={2020 IEEE/RSJ International Conference on Intelligent Robots and Systems (IROS)},
  pages={7447--7453},
  year={2020},
  organization={IEEE}
}

@article{lachner2020influence,
  title={The influence of coordinates in robotic manipulability analysis},
  author={Lachner, Johannes and Schettino, Vincenzo and Allmendinger, Felix and Fiore, Mario Daniele and Ficuciello, Fanny and Siciliano, Bruno and Stramigioli, Stefano},
  journal={Mechanism and machine theory},
  volume={146},
  pages={103722},
  year={2020},
  publisher={Elsevier}
}

@article{di2021effects,
  title={Effects of dynamic model errors in task-priority operational space control},
  author={Di Lillo, Paolo and Antonelli, Gianluca and Natale, Ciro},
  journal={Robotica},
  volume={39},
  number={9},
  pages={1642--1653},
  year={2021},
  publisher={Cambridge University Press}
}

@article{di2021framework,
  title={A framework for set-based kinematic control of multi-robot systems},
  author={Di Lillo, Paolo and Pierri, Francesco and Antonelli, Gianluca and Caccavale, Fabrizio and Ollero, Anibal},
  journal={Control Engineering Practice},
  volume={106},
  pages={104669},
  year={2021},
  publisher={Elsevier}
}

@article{lachner2021energy,
  title={Energy budgets for coordinate invariant robot control in physical human--robot interaction},
  author={Lachner, Johannes and Allmendinger, Felix and Hobert, Eddo and Hogan, Neville and Stramigioli, Stefano},
  journal={The International Journal of Robotics Research},
  volume={40},
  number={8-9},
  pages={968--985},
  year={2021},
  publisher={Sage Publications Sage UK: London, England}
}

@article{kazemipour2021motion,
  title={Motion Control of Redundant Robots with Generalised Inequality Constraints},
  author={Kazemipour, Amirhossein and Khatib, Maram and Khudir, Khaled Al and De Luca, Alessandro},
  journal={arXiv preprint arXiv:2110.01689},
  year={2021}
}

@INPROCEEDINGS{osorio2022,
  author={Osorio, Juan D. Muñoz and Allmendinger, Felix},
  booktitle={2022 IEEE International Conference on Autonomous Robot Systems and Competitions (ICARSC)}, 
  title={A Suitable Hierarchical Framework with Arbitrary Task Dimensions under Unilateral Constraints for physical Human Robot Interaction}, 
  year={2022},
  volume={},
  number={},
  pages={66-72},
  doi={10.1109/ICARSC55462.2022.9784782}}

\end{document}